\documentclass{easychair}

\usepackage{doc}

\usepackage{pifont}
\usepackage{xcolor}
\definecolor{second}{HTML}{007000}

\usepackage{comment}

\usepackage{amssymb}
\usepackage{pifont}
\usepackage{booktabs}
\usepackage{longtable}
\usepackage{pgfplotstable}
\usepackage{titlesec}
\usepackage{multirow}
\usepackage{subcaption}
\usepackage{fancyvrb}
\usepackage{cleveref}
\usepackage{threeparttable}
\usepackage{wrapfig}
\usepackage{makecell}

\pgfplotsset{compat=1.16}

\newcommand{\xmark}{\ding{55}}

\pgfkeysifdefined{/pgfplots/table/output empty row/.@cmd}{
    \pgfplotstableset{
        empty header/.style={
          every head row/.style={output empty row},
        }
    }
}{
    \pgfplotstableset{
        empty header/.style={
            typeset cell/.append code={%
                \ifnum\pgfplotstablerow=-1 %
                    \pgfkeyssetvalue{/pgfplots/table/@cell content}{}%
                \fi
            }
        }
    }
}

\titlespacing*{\paragraph}
{0pt}{2pt}{5pt}

\title{The Third International Verification of Neural Networks Competition (VNN-COMP 2022): Summary and Results}%

\author{Mark Niklas M\"uller\footnotemark[1]\footnotetext{\footnotemark[1]$\;\;$Equal contribution}${\;\,}$\inst{1}
        \and Christopher Brix\footnotemark[1]${\;\,}$\inst{2}
        \and Stanley Bak\inst{3}
        \and Changliu Liu\inst{4}
        \and Taylor T. Johnson\inst{5}
}

\institute{
    ETH Zurich, Zurich, Switzerland\\
    \email{mark.mueller@inf.ethz.ch}
    \and
    RWTH Aachen University, Aachen, Germany\\
    \email{brix@cs.rwth-aachen.de}
    \and
    Stony Brook University, Stony Brook, New York, USA\\
    \email{stanley.bak@stonybrook.edu}
    \and
    Carnegie Mellon University, Pittsburgh, Pennsylvania, USA\\
    \email{cliu6@andrew.cmu.edu}
    \and
    Vanderbilt University, Nashville, Tennessee, USA\\
    \email{taylor.johnson@vanderbilt.edu}
 }

\authorrunning{M. M\"uller, C. Brix, S. Bak, C. Liu, T. Johnson}
\titlerunning{VNN-COMP 2022 Report}

\date{\phantom{date}}

\begin{document}

\maketitle

\begin{abstract}
This report summarizes the 3rd International Verification of Neural Networks Competition (VNN-COMP 2022), held as a part of the 5th Workshop on Formal Methods for ML-Enabled Autonomous Systems (FoMLAS), which was collocated with the 34th International Conference on Computer-Aided Verification (CAV). 
VNN-COMP is held annually to facilitate the fair and objective comparison of state-of-the-art neural network verification tools, encourage the standardization of tool interfaces, and bring together the neural network verification community.
To this end, standardized formats for networks (ONNX) and specification (VNN-LIB) were defined, tools were evaluated on equal-cost hardware (using an automatic evaluation pipeline based on AWS instances), and tool parameters were chosen by the participants before the final test sets were made public.
In the 2022 iteration, 11 teams participated on a diverse set of 12 scored benchmarks.
This report summarizes the rules, benchmarks, participating tools, results, and lessons learned from this iteration of this competition.
\end{abstract}




\section{Introduction}
\label{sec:introduction}

Deep Learning based systems are increasingly being deployed in a wide range of domains, including recommendation systems, computer vision, and autonomous driving. While the nominal performance of these methods has increased significantly over the last years, often even surpassing human performance, they largely lack formal guarantees on their behavior. 
However, in safety-critical applications, including autonomous systems, robotics, cybersecurity, and cyber-physical systems, such guarantees are essential for certification and confident deployment. 

While the literature on the verification of traditionally designed systems is wide and successful, neural network verification remains an open problem, despite significant efforts over the last years. In 2020, the International Verification of Neural Networks Competition (VNN-COMP) was established to facilitate comparison between existing approaches, bring researchers working on this problem together, and help shape future directions of the field. In 2022, the 3rd and most recent iteration of the annual VNN-COMP\footnote{\url{https://sites.google.com/view/vnn2022/home}} was held as a part of the 5th Workshop on Formal Methods for ML-Enabled Autonomous Systems (FoMLAS) that was collocated with the 34th International Conference on Computer-Aided Verification (CAV). 

This third iteration of the VNN-COMP continues last year's trend of increasing standardization and automatization, aiming to enable a fair comparison between the participating tool and simplify the evaluation of a large number of tools on a variety of (real-world) problems. 
The VNN-COMP 2022 standardizes 1) neural network and specification formats, ONNX for neural networks and VNN-LIB for specifications, 2) evaluation hardware, providing participants the choice of a range of cost-equivalent AWS instances with different trade-offs between CPU and GPU performance, and 3) evaluation pipelines, enforcing a uniform interface for the installation and evaluation of tools. 

The competition was kicked off with the solicitation for participation in February 2022. By March, several teams had registered, allowing the rule discussion to be finalized in April 2022 (see an overview in \Cref{sec:rules}). From April to June 2022, benchmarks were proposed and discussed. Meanwhile, the organizing team decided to continue using AWS as the evaluation platform and started to implement an automated submission and testing system for both benchmarks and tools. By mid-July 2022, eleven teams submitted their tools and the organizers evaluated all entrants to obtain the final results, discussed in \Cref{sec:results} and presented at FoMLAS on July 31st, 2022. 
Discussions were structured into three issues on the official GitHub repository\footnote{\url{https://github.com/stanleybak/vnncomp2022/issues}}: rules discussion, benchmarks discussion, and tool submission. All submitted benchmarks\footnote{\url{https://github.com/ChristopherBrix/vnncomp2022_benchmarks}} and final results\footnote{\url{https://github.com/ChristopherBrix/vnncomp2022_results}} were aggregated in separate GitHub repositories. 

The remainder of this report is organized as follows: \Cref{sec:rules} discusses the competition rules, \Cref{sec:participants} lists all participating tools, \Cref{sec:benchmarks} lists all benchmarks,  \Cref{sec:results} summarizes the results, and \Cref{sec:conclusion} concludes the report, discussing potential future improvements.

\newpage

\section{Rules}
\label{sec:rules}

\paragraph*{Terminology}
An \emph{instance} is defined by a property specification (pre- and post-condition), a network, and a timeout). 
For example, one instance might consist of an MNIST classifier with one input image, a given local robustness threshold $\epsilon$, and a specific timeout.
A \emph{benchmark} is defined as a set of related instances.
For example, one benchmark might consist of a specific MNIST classifier with 100 input images, potentially different robustness thresholds $\epsilon$, and one timeout per input.

\paragraph*{Run-time caps}
Run-times are capped on a per-instance basis, i.e., any verification instance will timeout (and be terminated) after at most X seconds, determined by the benchmark proposer. These can be different for each instance. 
The total per-benchmark runtime (sum of all per-instance timeouts) may not exceed 6 hours per benchmark. 
For example, a benchmark proposal could have six instances with a one-hour timeout, or 100 instances with a 3.6-minute timeout, each.
To enable a fair comparison, we measure the startup overhead for each tool by running it on a range of tiny networks and subtract the minimal overhead from the total runtime.

\paragraph*{Hardware}
To allow for comparability of results, all tools were evaluated on equal-cost hardware using  Amazon Web Services (AWS).
Each team could decide between a range of AWS instance types (see \Cref{tab:instances}) providing a CPU, GPU, or mixed focus.
Except for the much weaker t2.large instance, all instances were priced at around three dollars per hour.

\begin{table}[h]
\centering
\caption{Available AWS instances.}\label{tab:instances}
\renewcommand{\arraystretch}{1.1}
\scalebox{0.985}{
\begin{tabular}{lccc} \toprule
         &vCPUs & RAM [GB] & GPU \\ 
         \midrule
         r5.12xlarge & 48 & 384 & \xmark \\ 
         p3.2xlarge  & 8 & 61 & V100 GPU with 16 GB memory \\
         m5.16xlarge  & 64 & 256 & \xmark \\
         g5.8xlarge  & 32 & 128 & A10G GPU with 24 GB memory \\
         t2.large  & 2 & 8 & \xmark \\
         \bottomrule
\end{tabular}
}
\end{table}

\paragraph{Scoring} 
The final score is aggregate as the sum of all benchmark scores.
Each benchmark score is the number of points (sum of instance scores discussed below) achieved by a given tool, normalized by the maximum number of points achieved by any tool on that benchmark. 
Thus, the tool with the highest sum of instance scores for a benchmark will get a benchmark score of 100, ensuring that all benchmarks are weighted equally, regardless of the number of constituting instances.

\paragraph*{Instance score}
Each instance is scored is as follows: 
\begin{itemize}\setlength{\itemsep}{0pt}
    \item Correct hold (property proven): 10 points;
    \item Correct violated (counterexample found): 10 points;
    \item Incorrect result: -100 points.
\end{itemize}
However, the ground truth for any given instance is generally not known a priori. In the case of disagreement between tools, we, therefore, place the burden of proof on the tool claiming that a specification is violated, i.e. that a counter-example can be found, and deem it correct exactly if it produces a valid counter-example.

\paragraph*{Time bonus}
Time bonuses are computed as follows.
\begin{itemize}\setlength{\itemsep}{0pt}
    \item The fastest tool for each solved instance will receive +2 points. 
    \item The second fastest tool will receive +1 point. 
\end{itemize}

All runtimes below 1.0 seconds after overhead correction (explained below) are considered to be 1.0 seconds exactly for scoring purposes. If two tools have runtimes within 0.2 seconds (after all corrections), for scoring purposes we will consider them the same runtime. Thus, multiple tools can receive the bonus for the fastest runtime.

\paragraph{Overhead Correction}
According to the rules discussion, we decided to subtract tool overhead time from the measured runtimes. For example, simply importing TensorFlow from Python and acquiring a GPU can sometimes take about 5 seconds. For particularly easy benchmarks where verification times can be significantly below one second, this would dominate the runtime. However, in a real-world application, typically, a large number of instances are analyzed without terminating the tool in between, leading to this overhead only being incurred once. Thus, we believe these corrected times to be more representative of real-world runtime behavior.

To measure the tool-specific overhead, we created trivial network instances and included those in the measurements. We then observed the minimum verification time over all instances and considered that to be the overhead time for the tool.

\paragraph{Format}
This year we standardized neural networks to be in \texttt{onnx} format, specifications in \texttt{vnnlib} format, and counter-examples in a new format similar to the \texttt{vnnlib} format.
Further, tool authors were required to provide scripts fully automating the installation process of their tool, including the acquisition of any licenses that might be needed. Similar to the previous year, a preparation and execution script had to be provided for running their tool on a specific instance consisting of a network file, specification file, and timeout.
The specifications are interpreted as definitions of counter-examples, meaning that a property is proven ``correct'' if the specification is shown to be unsatisfiable, conversely, the property is shown to be violated if a counterexample fulfilling the specification is found. 
Specifications consisted of disjunctions over conjunctions in both pre- and post-conditions, allowing a wide range of properties from adversarial robustness over multiple hyper-boxes to safety constraints to be encoded.
For example, robustness with respect to inputs in a hyper-box had to be encoded as disjunctive property, where any of the other classes is predicted.

\newpage

\section{Participants}
\label{sec:participants}
We list the tools and teams that participated in the VNN-COMP 2022 in \Cref{tab:tools} and reproduce their own descriptions of their tools below.


\begin{table*}[h]
\begin{center}
\begin{minipage}{\linewidth}
\caption{Summary of the key features of participating tools. The hardware column describes the used AWS instance with \texttt{p3} and \texttt{g5} making GPUs available, see \Cref{tab:instances} for more details. Licenses refer to the external licenses required to use the corresponding tool, not the licensing of the tool itself.}%
\label{tab:tools}%
\renewcommand{\arraystretch}{1.25}
\resizebox{\textwidth}{!}{
\begin{tabular}{lcm{5.0cm}ccc} \toprule
         Tool & References & Organizations & Place & Hardware & Licenses\\
         \midrule
         \textbf{$\boldsymbol{\alpha}$,$\boldsymbol{\beta}$-CROWN} & \cite{xu2020automatic,xu2021fast,wang2021betacrown,zhang2022general} & CMU, Drexel, UCLA, UIUC, Columbia & \textbf{1} & \texttt{g5} & GUROBI, IBM CPLEX  \\
         AveriNN & \cite{prabhakar2019abstraction} & Kansas State University & 11 & t2 & - \\
         CGDTest & - & University of Waterloo & 5 & m5 & -\\
         Debona & \cite{brix2020debona} & RWTH Aachen University & 8 & m5 & GUROBI \\
         FastBATLLNN & \cite{FerlezKS22} & University of California & 9 & t2 & -\\
         Marabou & \cite{KatzHIJLLSTWZDK19} & Hebrew University of Jerusalem, Stanford University, NRI Secure & 7 & m5 & GUROBI\\
         \textbf{\textsc{MN-BaB}}& \cite{mnbab,singh2019krelu,mueller2021prima,DeepPoly:19} & ETH Zurich & \textbf{2} & \texttt{p3} & GUROBI\\
         nnenum & \cite{bak2021nnenum} & Stony Brook University & 4 & m5 & - \\
         PeregriNN & \cite{khedr2021peregrinn} & University of California & 6 & m5 & GUROBI\\
         VeraPak & \cite{Smith2021} & Utah State University & 10 & t2 & - \\
         \textbf{VeriNet} & \cite{HenriksenLomuscio20, HenriksenLomuscio21, Henriksen+21} & Imperial College London & \textbf{3} & g5 & Xpress \\
         \bottomrule
\end{tabular}
}
\end{minipage}
\end{center}
\end{table*}


\subsection{$\alpha,\!\beta$-CROWN}

\paragraph*{Team} $\alpha,\!\beta$-CROWN is developed by a multi-institutional team from CMU, UCLA, Drexel University, UIUC, and Columbia University:
\begin{itemize}
    \item Developers: Huan Zhang (CMU), Kaidi Xu (Drexel), Zhouxing Shi (UCLA), Jinqi Chen (CMU), Linyi Li (UIUC), Shiqi Wang (Columbia), Zhuolin Yang (UIUC), Yihan Wang (UCLA)
    \item Advisors: Zico Kolter (CMU), Cho-Jui Hsieh (UCLA), Bo Li (UIUC), Suman Jana (Columbia)
\end{itemize}
\paragraph*{Description} $\alpha,\!\beta$-CROWN (\texttt{alpha-beta-CROWN}) is an efficient neural network verifier based on the linear bound propagation framework, built on a series of works on bound-propagation-based neural network verifiers:  CROWN~\cite{zhang2018efficient,xu2020automatic}, $\alpha$-CROWN~\cite{xu2021fast}, $\beta$-CROWN~\cite{wang2021betacrown} and GCP-CROWN~\cite{zhang2022general}. The original version of CROWN was also concurrently proposed as DeepPoly~\cite{DeepPoly:19} (mathematically equivalent). Our latest work of linear bound propagation with general cutting planes (GCP-CROWN~\cite{zhang2022general}) is currently the most general formulation of the linear bound propagation framework for neural network verification.

We use the generalized version of bound propagation algorithms in the \texttt{auto\_LiRPA} library~\cite{xu2020automatic} which supports general neural network architectures (including convolutional layers, pooling layers, residual connections, recurrent neural networks, and Transformers) and a wide range of activation functions (e.g., ReLU, tanh, sigmoid,  max pooling and average pooling), and is efficiently implemented on GPUs with Pytorch and CUDA. We jointly optimize intermediate layer bounds and final layer bounds using gradient ascent (referred to as $\alpha$-CROWN or optimized CROWN/LiRPA~\cite{xu2021fast}). Most importantly, we use branch and bound~\cite{bunelunified2018} (BaB) and incorporate split constraints in BaB into the bound propagation procedure efficiently via the $\beta$-CROWN algorithm~\cite{wang2021betacrown}, and use cutting-plane method in GCP-CROWN~\cite{zhang2022general} to further tighten the bound.
For smaller networks, we also use a mixed integer programming (MIP) formulation~\cite{Tjeng2019EvaluatingRO} combined with tight intermediate layer bounds from $\alpha$-CROWN (referred to as $\alpha$-CROWN + MIP~\cite{zhang2022general}). The combination of efficient, optimizable and GPU-accelerated bound propagation with BaB produces a powerful and scalable neural network verifier.

\paragraph*{Link} \url{https://github.com/Verified-Intelligence/alpha-beta-CROWN} (latest version)
\paragraph*{Competition submission} \url{https://github.com/huanzhang12/alpha-beta-CROWN_vnncomp22} (only for reproducing competition results; please use the latest version for other purposes)
\paragraph*{Hardware and licenses} CPU and GPU with 32-bit or 64-bit floating point; Gurobi license required for the \texttt{mnistfc} benchmark; IBM CPLEX required for the \texttt{oval21} benchmark.
\paragraph*{Participated benchmarks} All benchmarks.

\subsection{AVeriNN}
\paragraph*{Team} Vishnu Bondalakunta (Kansas State University), Pavithra Prabhakar (Kansas State University)

\paragraph*{Description} The Abstraction Verification of Neural Networks Tool (AVeriNN) is written in Python 3 and performs reachability analysis of abstract systems called Interval Neural Networks \cite{prabhakar2019abstraction}. Potentially large-scale neural networks are converted to smaller Interval Neural Networks by clustering similar neurons together. The degree of reduction of the neural network is user-controlled via the 'delta' parameter. AVeriNN uses geometric representations that allow for a layer-by-layer computation of the reachable set of Interval Neural Networks, similar to NNV \cite{tran2019fm}. However, the difference is that instead of using star sets, an extended version of star sets named Interval star sets are used. Furthermore, the obtained reach set is always an over-approximation of the exact reachable set of the original neural network.

The current implementation is a baseline proof-of-concept devoid of optimizations present in other star set based tools such as NNV and nnenum.

\paragraph*{Link}
\url{https://github.com/vishnuteja97/AVeriNN.git}

\paragraph*{Commit}
38cb39df826b76bb0c396faf0e13a24c02e4afad

\paragraph*{Hardware and licences}
CPU, No licenses required \nopagebreak
\nopagebreak
\paragraph*{Participated benchmarks}  \texttt{rl\_benchmarks}

\subsection{CGDTest}
\paragraph{Team} Vineel Nagisetty (Borealis AI), Guanting Pan (University of Waterloo), Piyush Jha (University of Waterloo), Christopher Srinivasa (Borealis AI), Vijay Ganesh (University of Waterloo)

\paragraph{Description} CGDTest is a DNN testing algorithm, which takes a DNN model,
constraints, and label as inputs; it aims to find an input such that the constraint needs
to be satisfied by the model. CGDTest is best viewed as a gradient-descent optimization 
method: Firstly, CGDTest can parse user-specified constraints and convert them into a 
differentiable constraint loss function. When an input is constructed to satisfy all the 
constraints during iterations, the constraint loss function is 0 for this input. Secondly, 
starting from a random input, CGDTest uses gradient descent to modify it to meet the 
stopping criteria. The stopping criteria are met when the constraint loss function results 
in 0.

\paragraph{Link} \url{https://github.com/vin-nag/CGD.git}
\paragraph{Commit} 57baf10076968f901fcab4751c530728c94e86a4
\paragraph{Hardware and licenses} CPU and GPU; No licenses required
\paragraph{Participated benchmarks} All benchmarks

\subsection{Debona}

\paragraph*{Team}
Christopher Brix (RWTH Aachen University), Thomas Noll (RWTH Aachen University)

\paragraph*{Description}
Debona \cite{brix2020debona} is a verification toolkit built upon the 2020 version of VeriNet \cite{HenriksenLomuscio20}.
In addition to the so-called Error-based Symbolic Interval Propagation (ESIP), which relies on parallel upper and lower bounds for each neuron, it also supports Reversed
Symbolic Interval Propagation (RSIP) with independent lower and upper bounds \cite{DeepPoly:19,zhang2018efficient}.
As RSIP is computationally more expensive, Debona first tries to verify the given property using ESIP, and falls back to RSIP only if this first attempt fails.

\paragraph*{Link}
\url{https://github.com/ChristopherBrix/Debona}

\paragraph*{Commit}
675d5c450960a7c341ea1f42f9d11f918f9e1417

\paragraph*{Hardware and licences}
CPU, Gurobi license

\paragraph*{Participated benchmarks}
\texttt{mnist\_fc}, \texttt{nn4sys}, \texttt{reach\_prob\_density}, \texttt{rl\_benchmarks}, \\ \texttt{tllverifybench}

\subsection{FastBATLLNN}
\paragraph{Team} James Ferlez (Developer), Haitham Khedr (Tester), and Yasser Shoukry (Supervisor) (University of California, Irvine)
\paragraph{Description} FastBATLLNN \cite{FerlezKS22} is a fast verifier of box-like (hyper-rectangle) output properties for Two-Level Lattice (TLL) Neural Networks (NN). FastBATLLNN uses both the unique semantics of the TLL architecture and the decoupled nature of box-like output constraints to provide a fast, polynomial-time verification algorithm: that is, polynomial-time in the number of neurons in the TLL NN to be verified (for a fixed input dimension). FastBATLLNN fundamentally works by converting the TLL verification problem into a region enumeration problem for a hyperplane arrangement (the arrangement is jointly derived from the TLL NN and the verification property). However, as FastBATLLNN leverages the unique properties of TLL NNs and box-like output constraints, its use is necessarily limited to verification problems formulated in those terms. Hence, FastBATLLNN can only compete on the \texttt{tllverifybench} benchmark.
\paragraph{Link} \url{https://github.com/jferlez/FastBATLLNN-VNNCOMP}
\paragraph*{Commit}
6100258b50a3fadf9792aec4ccf39ed14778e338
\paragraph{Hardware and licenses} CPU; no licenses required
\paragraph{Participated benchmarks} \texttt{tllverifybench}

\subsection{Marabou}
\paragraph{Team} Haoze Wu (Stanford University), Aleksandar Zeljic (Stanford University), Teruhiro Tagomori (Stanford University/NRI Secure),  Clark Barrett (Stanford University), Guy Katz (Hebrew University of Jerusalem)
\paragraph{Description}  Marabou~\cite{katz2019marabou} is a user-friendly Neural Network Verification toolkit that can answer queries about a network’s properties by encoding and solving these queries as constraint satisfaction problems. It has both Python/C++ APIs through which users can load neural networks and define arbitrary linear properties over the neural network. Marabou supports many different linear, piecewise-linear, and non-linear~\cite{wu2022toward} operations and architectures (e.g., FFNNs, CNNs, residual connections, Graph Neural Networks~\cite{vegas}). 

Under the hood, Marabou employs a uniform solving strategy for a given verification query. In particular, Marabou performs complete analysis that employs a specialized convex optimization procedure~\cite{wu2022efficient} and abstract interpretation~\cite{DeepPoly:19,vegas}. It also uses the Split-and-Conquer algorithm~\cite{wu2020parallelization} for parallelization.

\paragraph{Link} \url{https://github.com/NeuralNetworkVerification/Marabou}
\paragraph*{Commit} 48a6f68267ee9e96dae7751f902139eac868fff3
\paragraph{Hardware and Licenses} CPU, no license required. Can also be accelerated with Gurobi (which requires a license)
\paragraph{Participated benchmarks} \texttt{cifar100\_tinyimagenet\_resnet}, \texttt{reach\_prob\_density}, \texttt{mnist\_fc}, \texttt{oval21}, \texttt{rl\_benchmarks}

\subsection{MN-BaB}

\paragraph*{Team}
Mark Niklas Müller (ETH Zurich), Robin Staab (ETH Zurich), Timon Gehr (ETH Zurich), Claudio Ferrari (ETH Zurich), Martin Vechev (ETH Zurich)

\paragraph*{Description}
MN-BaB~\cite{mnbab} is an open-source neural network verifier leveraging precise multi-neuron constraints~\cite{singh2019krelu, mueller2021prima} combined with efficient GPU-enabled linear bound-propagation~\cite{DeepPoly:19} in a branch and bound framework. MN-BaB is complete for piece-wise linear activation functions and can handle fully-connected, convolutional, and residual network architectures containing ReLU, Sigmoid, Tanh, and Maxpool non-linearities. It uses either 64- or 32-bit precision depending on the benchmark. Multi-neuron relaxations of non-linear activations~\cite{mueller2021prima} and potential MILP-based refinement of individual neuron-bounds\cite{singh2019refinement} are computed on CPU, while the core bound propagation is fully run on GPU. 
Depending on the benchmark, MN-BaB can use different verification modes besides the standard linear bound propagation based BaB including input-domain splitting with forward bound propagation and full MILP encodings~\cite{TjengXT19}. These modes were selected manually per benchmark/network type.
MN-BaB is implemented using PyTorch~\cite{PaszkeGMLBCKLGA19}, and uses Gurobi~\cite{gurobi} for solving MILP instances. 

\paragraph*{Link}
\url{https://github.com/eth-sri/mn-bab}

\paragraph*{Commit}
Please use the up-to-date main repository for anything but reproducing VNN-COMP results.
\url{https://github.com/mnmueller/mn_bab_vnn_2022}\\ (c67bcd19e9e2cff13f8c2785c94034b89b47ee05)

\paragraph*{Hardware and licences}
CPU and GPU, GUROBI License

\paragraph*{Participated benchmarks} All benchmarks.

\subsection{nnenum}
\paragraph*{Team} Stanley Bak (Stony Brook University)

\paragraph*{Description} 
The nnenum tool~\cite{bak2021nnenum} uses multiple levels of abstraction to achieve high-performance verification of ReLU networks without sacrificing completeness~\cite{bak2020vnn}. Analysis combines three types of zonotopes with star set (triangle) overapproximations~\cite{tran2019fm}, and uses efficient parallelized ReLU case splitting~\cite{bak2020cav}. 
The ImageStar method~\cite{tran2020cav} allows sets to be quickly propagated through all layers supported by the ONNX runtime, such as convolutional layers with arbitrary parameters.  
The tool is written in Python 3 and uses GLPK for LP solving.
New this year we used code from DNNV~\cite{shriver-etal:CAV:2021:dnnv} in order to convert maxpooling layers into multiple ReLU layers, which permitted analysis of more networks including VGGNET.

\paragraph*{Link}
\url{https://github.com/stanleybak/nnenum}

\paragraph*{Commit}
c93a39cb568f58a26015bd151acafab34d2d4929

\paragraph*{Hardware and licences}
CPU, No licenses required

\paragraph*{Participated benchmarks}
\texttt{AcasXu}, \texttt{cifar2020}, \texttt{mnist\_fc}, \texttt{oval21}.

\subsection{PeregriNN}
\paragraph*{Team} Haitham Khedr, James Ferlez, and Yasser Shoukry (University of California, Irvine)
\paragraph*{Description}

PeregriNN\cite{khedr2021peregrinn} is a sound and complete Neural Network verification tool written in Python 3 focusing on ReLU activations. It uses search and optimization to verify the NN property. PeregriNN uses symbolic interval analysis for bound approximation and a linear program to check for violations of the specification. The solution of the linear program is used as a heuristic to guide the neuron conditioning (fixing ReLUs) during abstraction refinement. Gurobi is used for LP solving and the framework this year adds support for ONNX networks and vnnlib specifications.

\paragraph*{Link}
\url{https://github.com/haithamkhedr/PeregriNN/tree/vnn2022}

\paragraph*{Commit}
4d315135dd34da5fc25b258fc4d40e1cf5a14ed9

\paragraph*{Hardware and licences}
CPU, Gurobi Academic license

\paragraph*{Participated benchmarks}
\texttt{collins\_rul\_cnn}, \texttt{mnist\_fc}, \texttt{nn4sys}, \texttt{oval21}, \texttt{reach\_prob\_density}, \texttt{rl\_benchmarks}, \texttt{tllverifybench}.

\newcommand{\verapak}{\textsc{VeRAPAk}}
\newcommand{\Verapak}{\textit{\textsc{Ve}rification for \textsc{R}obust neural networks through \textsc{A}bstraction, \textsc{P}artitioning and \textsc{A}ttac\textsc{k} methods}}
\subsection{V{\small E}RAPA{\small K}}

\paragraph{Team} Bennett DenBleyker$^1$, Joshua Smith$^2$, Viswanathan Swaminathan$^3$, Zhen Zhang$^1$.

$^1$Utah State University, $^2$Campbell Scientific Inc., $^3$Adobe Systems Inc.

\paragraph{Description} \Verapak\ (\verapak) is a neural network verification framework that, by combining various input region abstraction, verification, and partitioning strategies, can potentially serve any desired use-case~\cite{Smith2021}.
In abstraction, \verapak\ finds representative points for a given region and tests them to ensure safety. In verification, \verapak\ checks the region as a whole for safety. If verification fails or its result is unknown, the region is partitioned into multiple subregions for further abstraction and verification.
\verapak\ is built to handle a myriad of different use-cases, where verification strategies may range from checking every discrete point for safety or assuming safety after a certain threshold, to integrating another DNN verification tool entirely, or even focusing purely on generating counterexamples instead.

The version of \verapak\ used in VNNCOMP was the same that is typically used for model training: built to rapidly produce as many counterexamples as possible in a uniform distribution over the input region. These are then typically used for retraining to reinforce robustness against counterexamples.
The only changes for the competition were to use discrete search for verification and halt after the first iteration where it finds a counterexample. Even so, it tends to find around 20 to 50 counterexamples before halting.

For the purposes of VNNCOMP, \verapak\ used per-benchmark tuning to set granularity. For the networks we deemed to be image-based classification networks, we set $\frac{1}{256}$ as our granularity. For all others, we used $\frac{1}{1000}x$, where $x$ is the range of each dimension as a vector. Theoretically, any property that can be set via config file or command line flags may also be set per-benchmark. \verapak\ is also not floating point sound.

\paragraph{Link} \url{https://github.com/formal-verification-research/VERAPAK}

\paragraph{Commit} 9181ef5a40672c0d82e5405037c9d8911587ab0d

\paragraph{Hardware and licenses} CPU, No licenses required

\paragraph{Participated benchmarks}
\texttt{cifar\_biasfield},
\texttt{mnist\_fc},
\texttt{rl\_benchmarks},

\subsection{VeriNet}
\paragraph{Team} 
Patrick Henriksen, Alessio Lomuscio (Imperial College London).

\paragraph{Description} VeriNet~\cite{HenriksenLomuscio20, HenriksenLomuscio21, Henriksen+21} is a complete Symbolic Interval Propagation (SIP) based verification toolkit for feed-forward neural networks. The underlying algorithm utilizes SIP to create a linear abstraction of the network, which, in turn, is used in an LP-encoding to analyze the verification problem. A branch and bound-based refinement phase is used to achieve completeness. 

VeriNet implements various optimizations, including a gradient-based local search for counterexamples, optimal relaxations for Sigmoids, adaptive node splitting~\cite{HenriksenLomuscio20}, succinct LP-encodings, and a novel splitting heuristic that takes into account indirect effects splits have on succeeding relaxations~\cite{HenriksenLomuscio21}.

VeriNet supports a wide range of layers and activation functions, including Relu, Sigmoid, Tanh, fully connected, convolutional, max and average pooling, batch normalization, reshape, crop, and transpose operations, as well as additive residual connections.

Note that VeriNet subsumes the Deepsplit method presented in \cite{HenriksenLomuscio21}.

\paragraph*{Hardware and licences}
CPU and GPU, Xpress Solver license required for large networks. 

\paragraph*{Link} \url{https://github.com/vas-group-imperial/VeriNet} \newline (66b70f6fef311c721a772c78573bbfae1644bc5e). \newline 
Note: This is an old version of VeriNet; \emph{the version used for the competition is not public}. 

\paragraph*{Participated benchmarks} All benchmarks.

\newpage
\section{Benchmarks}
\label{sec:benchmarks}

In this section we provide an overview of all benchmarks, reproducing the benchmark proposers' descriptions.

\begin{table}[h]
    \centering
    \caption{Overview of all scored benchmarks. }
    \label{tab:my_label}
    \resizebox{\textwidth}{!}{
    \renewcommand{\arraystretch}{1.4}
    \begin{tabular}{cccccc}
    \toprule
    Category &
    Benchmark &
    Application &
    Network Types &
    \# Neurons &
    Effective Input Dim
    \\
    \midrule
    \multirow{2.6}{*}{Complex} & Carvana UNet  & Image Segmentation                            & Complex UNet              & 275k - 373k & 4.3k \\
                             & NN4Sys        & \makecell{Dataset Indexing \\ \&  Cardinality Prediction}   & Complex (ReLU + Sigmoid)  & 384 - 94k   & 1-4 \\
    \cmidrule(lr){1-6}
    \multirow{6}{*}{\makecell{CNN \\ \& ResNet}} & Cifar Bias Field      & Image Classification & Conv. + ReLU  & 45k   & 16 \\
                            & Collins RUL CNN       & Condition Based Maintenance               & Conv. + ReLU  & 5.5k - 28k   & 2 - 200 \\                            
                            & Large ResNets         & Image Classification                      & ResNet (Conv. + ReLU)  & 55k - 286k   & 3.1k - 12k \\
                            & Oval21                & Image Classification                      & Conv. + ReLU  & 3.1k - 6.2k   & 3.1k \\
                            & SRI ResNet A/B        & Image Classification                      & ResNet (Conv. + ReLU)  & 11k   & 3.1k \\
                            & VGGNet16              & Image Classification                      & Conv. + ReLU + MaxPool    & 13.6M & 1 - 95k \\
    \cmidrule(lr){1-6}
    \multirow{4}{*}{\makecell{FC}} & MNIST FC              & Image Classification                     & FC. + ReLU    & 512 - 1.5k & 784 \\
                            & Reach Prob Density    & Probability density estimation           & FC. + ReLU    & 64 - 192 & 3 - 14 \\
                            & RL Benchmarks  & Reinforcement Learning                   & FC. + ReLU    & 128 - 512  & 4 - 8 \\
                            & TLL Verify Bench      & Two-Level Lattice NN                     & \makecell{Two-Level Lattice NN \\(FC. + ReLU)}  & 252 - 16k & 2 \\
    \bottomrule
    \end{tabular}
    }
\end{table}

\subsection{Carvana UNet}
\paragraph{Proposed by} Yonggang Luo (Chongqing Changan Automobile Co. Ltd.), Jinyan Ma (Chongqing Changan Automobile Co. Ltd.).

\paragraph{Motivations} Currently, most Neural Network Verification benchmarks are focusing on image classification tasks. However, in many practical scenarios, e.g., autonomous driving, people may pay more attention to object detection or semantic segmentation~\cite{ronneberger2015u}  tasks. To bring some incentives for developing tools for more general purposes into the community, we proposed Carvana UNet for semantic segmentation tasks. We advocated that tools should handle more practical architectures and Carvana UNet is the first step towards this goal.

\paragraph*{Networks} The proposed Carvana UNet include two simplified UNet models. Model one is composed of four Conv2d layers which are followed by a BN and ReLu layer; for model two, we add one AveragePool layer and one TransposedConv Upsampling layer~\cite{tran2021robustness} to model one. The input size is [1, 4, 31, 47], where 1 is the batchsize, 4 is the number of channels, and 31 and 47 are the height and the width of samples respectively. The first three channels represent RGB values of images. The last channel represents the model-produced mask, which is used for calculating the number of correctly predicted pixels by the model. The model has one output, which is the number of correctly predicted pixels by the model, compared with the model-produced mask.

\paragraph*{Specifications} The Carvana dataset consists of 5088 images covering 318 cars, which means each car has 16 images. We choose one image for each car, 318 images in total, as a test set. And the remaining 4700 images are used for training. There are 52 images whose 98.8 percent pixels can be predicted correctly for model one, and 44 images whose 99 percent pixels can be predicted correctly for model two. We randomly select 16 images from these images for verification. For each image, we specify the property that the number of correctly predicted pixels by the model is always greater than 1314 (90 percent of the total pixels) within $\ell_\infty$ norm input perturbation of $\epsilon=\frac{1}{255}$ or  $\epsilon=\frac{3}{255}$. The per-example timeout is set to 300 seconds.

\paragraph*{Link} \url{https://github.com/pomodoromjy/vnn-comp-2022-Carvana-unet}

\subsection{NN4Sys}
\paragraph*{Proposed by} the $\alpha,\!\beta$-CROWN team with collaborations with Cheng Tan and Haoyu He at Northeastern University.
\paragraph*{Application}
The benchmark contains networks for database learned index and learned cardinality
estimation, which maps input from various dimensions to a single scalar as output. 

\begin{itemize}

\item \textit{Background}: learned index and learned cardinality are all instances in neural networks for computer systems (NN4Sys), which are neural network based methods performing system operations. These classes of methods show great potential but have one drawback---the outputs of an NN4Sys model (a neural network) can be arbitrary, which may lead to unexpected issues in systems.

\item \textit{What to verify}: our benchmark provides multiple pairs of (1) trained NN4Sys model
and (2) corresponding specifications. We design these pairs with different parameters such
that they cover a variety of user needs and have varied difficulties for verifiers. 
We describe benchmark details in our NN4SysBench report~\cite{he2022Characterizing}: \url{http://naizhengtan.github.io/doc/papers/characterizing22haoyu.pdf}.

\item \textit{Translating NN4Sys applications to a VNN benchmark}: 
the original NN4Sys applications have sophisticated features that are hard to express.
We tailored the neural networks and their specifications to be suitable for VNN-COMP.
For example, learned index~\cite{kraska18case} contains multiple NNs in a tree structure that together serve one purpose.
However, this cascading structure is inconvenient/unsupported to verify
because there is a ``switch" operation---choosing one NN in the second stage
based on the prediction of the first stage's NN.
To convert learned indexes to a standard form, we merge the NNs into one larger NN.

\item \textit{A note on broader impact}: using NNs for systems is a broad topic, but many existing works
lack strict safety guarantees. We believe that NN Verification can help system developers gain confidence
to apply NNs to critical systems. We hope our benchmark can be an early step toward this vision.

\end{itemize}

\paragraph*{Networks}
This benchmark has six networks with different parameters: two for learned indexes
and four for learned cardinality estimation.
The learned index uses fully-connected feed-forward neural networks.
The cardinality estimation has a relatively sophisticated internal structure;
please see our NN4SysBench report (URL listed above) for details.

\paragraph*{Specifications}
For learned indexes,
the specification aims to check if the prediction error is bounded.
The specification is a collection of pairs of input and output intervals such that
any input in the input interval should be mapped to the corresponding output interval.
For learned cardinality estimation,
the specifications check the prediction error bounds (similar to the learned indexes)
and monotonicity of the networks.
By monotonicity specifications, we mean that for two inputs, the network should produce a larger
output for the larger input, which is required by cardinality estimation.

\paragraph{Link:} \url{https://github.com/Cli212/VNNComp22_NN4Sys}

\pagebreak
\subsection{Cifar Bias Field}

\paragraph{Proposed by} The VeriNet team.

\paragraph*{Motivation} This benchmark considers verification of a Cifar 10 network against bias field perturbations. The bias field perturbations are encoded by creating augmented networks with only 16 input parameters; thus, the problem has a significantly lower input dimensionality than many other image-based benchmarks.

\paragraph*{Networks} For each image to be verified, a separate bias field transform network is created~\cite{Henriksen+21} which consists of the FC transform layer followed by the Cifar CNN. The Cifar CNN consists of 8 convolutional layers followed by ReLUs. Each bias field transform network has 363k parameters and 45k nodes.  

\paragraph*{Specifications} The specification considers bias field perturbations of the input. The task is reduced to a standard $\ell_\infty$ specification by encoding bias field the bias field transformation into fully connected layers which are prepended to the network under consideration. The bias field perturbations used $\epsilon = 0.06$ (as described in~\cite{Henriksen+21}) and a timeout of 5 minutes.

\subsection{Collins-RUL-CNN}
\paragraph*{Proposed by} Collins Aerospace, Applied Research and Technology.

\paragraph*{Motivation} Machine Learning (ML) is a disruptive technology for the aviation industry. This particularly concerns safety-critical aircraft functions, where high-assurance design and verification methods have to be used in order to obtain approval from certification authorities for the new ML-based products. Assessment of correctness and robustness of trained models, such as neural networks, is a crucial step for demonstrating the absence of unintended functionalities. The key motivation for providing this benchmark is to strengthen the interaction between the VNN community and the aerospace industry by providing a realistic use case for neural networks in future avionics systems.

\paragraph*{Application} Remaining Useful Life (RUL) is a widely used metric in Prognostics and Health Management (PHM) that manifests the remaining lifetime of a component (e.g., mechanical bearing, hydraulic pump, aircraft engine). RUL is used for Condition-Based Maintenance (CBM) to support aircraft maintenance and flight preparation. It contributes to such tasks as augmented manual inspection of components and scheduling of maintenance cycles for components, such as repair or replacement, thus moving from preventive maintenance to \emph{predictive} maintenance (do maintenance only when needed, based on component’s current condition and estimated future condition). This could allow to eliminate or extend service operations and inspection periods, optimize component servicing (e.g., lubricant replacement), generate inspection and maintenance schedules, and obtain significant cost savings. Finally, RUL function can also be used in airborne (in-flight) applications to dynamically inform pilots on the health state of aircraft components during flight. Multivariate time series data is often used as RUL function input, for example, measurements from a set of sensors monitoring the component state, taken at several subsequent time steps (within a time window). Additional inputs may include information about the current flight phase, mission, and environment. Such highly multi-dimensional input space motivates the use of Deep Learning (DL) solutions with their capabilities of performing automatic feature extraction from raw data.

\paragraph*{Networks} The benchmark includes 3 convolutional neural networks (CNNs) of different complexity: different numbers of filters and different sizes of the input space. All networks contain only convolutional and fully connected layers with ReLU activations. All CNNs perform the regression function. They have been trained on the same dataset (time series data for mechanical component degradation during flight).

\paragraph*{Specifications} We propose 3 properties for the NN-based RUL estimation function. First, two properties (robustness and monotonicity) are local, i.e., defined around a given point. We provide a script with an adjustable random seed that can generate these properties around input points randomly picked from a test dataset. For robustness properties, the input perturbation (delta) is varied between 5\% and 40\%, while the number of perturbed inputs varies between 2 and 16. For monotonicity properties, monotonic shifts between 5\% and 20\% from a given point are considered. Properties of the last type ("if-then") require the output (RUL) to be in an expected value range given certain input ranges. Several if-then properties of different complexity are provided (depending on range widths).

\paragraph*{Link} \url{https://github.com/loonwerks/vnncomp2022}

\subsection{Large ResNets}
\paragraph*{Proposed by} the $\alpha,\!\beta$-CROWN team.


\paragraph*{Motivations} With the rapid development of large ML models, it would be interesting to test the scalability of current verification methods. We aim to keep scaling up our model size upon last year's residual networks~\cite{he2016deep} (ResNet) structure and introducing more challenging datasets: TinyImageNet and CIFAR-100 with higher input dimension and more classification labels than most existing benchmarks. The purpose of this benchmark is to evaluate verification tools' scalability to large models.

\paragraph*{Networks} We provide one large ResNet model for TinyImageNet, and four ResNet models on CIFAR-100 with different model widths and depths. These models are trained with a combination of CROWN-IBP~\cite{zhang2019towards} and adversarial training~\cite{madry2017towards}. The smallest model is 11-layer, and the largest model has 21 layers. The largest model has 286,820 neurons. The models include:

\vspace{1em}
\noindent \textbf{TinyImageNet} (input dimension $64 \times 64 \times 3$, 200 classes)
\begin{itemize}
    \item \texttt{TinyImageNet-ResNet-medium}: 8 residual blocks, 17 convolutional layers + 2 linear layers
\end{itemize}

\vspace{1em}
\noindent \textbf{CIFAR-100 models} (input dimension $32 \times 32 \times 3$, 100 classes)
\begin{itemize}
    \item \texttt{CIFAR100-ResNet-small}: 4 residual blocks, 9 convolutional layers + 2 linear layers
    \item \texttt{CIFAR100-ResNet-medium}: 8 residual blocks, 17 convolutional layers + 2 linear layers
    \item \texttt{CIFAR100-ResNet-large}: 8 residual blocks, 19 convolutional layers + 2 linear layers (almost identical to standard ResNet-18 architecture)
    \item \texttt{CIFAR100-ResNet-super}: 9 residual blocks, 19 convolutional layers + 2 linear layers
\end{itemize}

The networks are trained via ``\textbf{mixed training}" by combining adversarial training loss and certified defense training (CROWN-IBP) loss with weights \textbf{0.05:0.95} for \texttt{CIFAR100-ResNet-small} and \texttt{CIFAR100-ResNet-medium}, and \textbf{0.01:0.99} for other models under $\ell_\infty$ perturbation norm of $\epsilon=\frac{1}{255}$. Empirically, we noticed that pure CROWN-IBP training results in low model accuracy and the models could be easily verified using IBP (not ideal for competition purposes), while pure adversarial training leads to unverifiable models. We found that a ``\textbf{mixed training}" by combining adversarial training and certified defense training loss could lead to an excellent balance between model clean accuracy and robustness, and is beneficial for obtaining higher verified accuracy. The model trained with mixed training cannot be directly verified by IBP, but their verified accuracy under a strong verifier such as $\alpha,\!\beta$-CROWN is higher than purely using CROWN-IBP training.

To gauge the verification hardness of these models, we evaluated all our trained models using a 100-step projected gradient descent (PGD) attack with 20 random restarts, and a simple bound propagation based verification algorithm CROWN~\cite{zhang2018efficient} (mathematically equivalent to the abstract interpretation used in DeepPoly~\cite{DeepPoly:19}) under $\ell_\infty$ norm perturbations. To ensure the appropriate level of difficulty, we use $\epsilon=\frac{1}{255}$ for all our models. The results are listed in Table~\ref{tab:resnet_model}.

\begin{table}[htb]
\centering
\scalebox{0.87}{
\renewcommand{\arraystretch}{1.2}
\begin{tabular}{cccccc}
\toprule
\multirow{3}{*}{Model} & \multirow{3}{*}{\# Parameters} & \multirow{3}{*}{\# ReLUs} & \multirow{3}{*}{Clean acc.}  & \multicolumn{2}{c}{$\epsilon=1/255$} \\
\cmidrule(lr){5-6}
                &       &                           &                             & \shortstack{PGD\\ Attack acc.}    & \shortstack{CROWN\\ Verified acc.}    \\ \midrule
\texttt{CIFAR100-ResNet-small}      & 5.4M        & 55\,460                      & 51.61\%                     & 37.92\%     & 20.14\%                \\
\texttt{CIFAR100-ResNet-medium}     & 10.1M         & 55\,460                     & 54.57\%                     & 40.42\%     & 29.08\%                \\
\texttt{CIFAR100-ResNet-large}    & 15.2M          & 286\,820                      & 53.24\%                     & 39.31\%     & 29.89\%                \\
\texttt{CIFAR100-ResNet-super}  & 31.6M            & 68\,836                     & 53.95\%                     & 38.83\%     & 27.53\%                \\
\cmidrule(lr){1-1}
\texttt{TinyImageNet-ResNet-medium}   & 14.4M           & 172\,296                     & 35.04\%                     & 21.85\%     & 13.51\%                \\
\bottomrule
\end{tabular}}
\caption{Clean accuracy, PGD accuracy, and CROWN verified accuracy for ResNet models. Note that the verified accuracy is obtained via the vanilla version of CROWN/DeepPoly, which has been widely used as a \emph{simple} baseline, not the $\alpha,\!\beta$-CROWN tool used in the competition.}
\label{tab:resnet_model}
\end{table}

\vbox{
\paragraph*{Specifications} We randomly select 10 images from the CIFAR-100 test set with a verification timeout of 240 seconds for the \texttt{CIFAR100-ResNet-small} model, 16 images with a timeout of 300 seconds for the \texttt{CIFAR100-ResNet-super} model and 24 images with a timeout of 200 seconds for other CIFAR-100 models. For TinyImageNet, we select 24 images from TinyImageNet test set with a timeout of 200 seconds for the \texttt{TinyImageNet-ResNet-medium} model. The overall runtime is guaranteed to be less than 6 hours. These images are \textbf{classified correctly} and cannot be attacked by a 100-steps PGD attack with 20 random restarts. We also filtered out the samples which can be verified by vanilla CROWN (which is used during training) to make the benchmark more challenging. The filtering process is done offline on a machine with a GPU due to the large sizes of these models.

\paragraph*{Link}\url{https://github.com/Lucas110550/CIFAR100_TinyImageNet_ResNet}
}

\subsection{Oval21}


\paragraph*{Proposed by} The OVAL team.

\paragraph*{Motivations}

The majority of adversarial robustness benchmarks consider image-independent perturbation radii, possibly resulting in some properties that are either easily verified by all verification methods, or too hard to be verified (for commonly employed timeouts) by any of them.
In line with the OVAL verification dataset from VNN-COMP 2020~\cite{vnn-comp}, whose versions have already been used in various recent works~\cite{bunel2020lagrangian,Lu2020Neural,depalma2021improved,depalma2021scaling,depalma2021sparsealgos,wang2021betacrown,xu2021fast,jaeckle2021generating, jaeckle2021neural}, the OVAL21 benchmark associates to each image-network pair a perturbation radius found via binary search to ensure that all properties are challenging to solve.

\paragraph*{Networks}
The benchmark includes 3 ReLU-based convolutional networks which were robustly trained~\cite{Wong2018} against $\ell_{\infty}$ perturbations of radius $\epsilon=2/255$ on CIFAR10.
Two of the networks, named \texttt{base} and \texttt{wide}, are composed of 2 convolutional layers followed by 2 fully connected layers and have respectively $3172$ and $6244$ activations. The third model, named \texttt{deep}, has 2 additional convolutional layers and a total of $6756$ activations.

\paragraph*{Specifications}
The verification properties represent untargeted adversarial
robustness (with respect to all possible misclassifications) to $\ell_{\infty}$ perturbations of varying $\epsilon$, with a per-instance timeout of $720$ seconds.
The property generation procedure relies on commonly employed lower and upper bounds to the adversarial loss to exclude perturbation radii that yield trivial properties.
10 correctly classified images per network are randomly sampled from the entire CIFAR10 test set, and a distinct $\epsilon \in [0, 16/255]$ is associated to each. First, a binary search is run to find the largest $\epsilon$ value for which a popular iterative adversarial attack \cite{dong2018boosting} fails to find an adversarial example. Then, a second binary search is run to find the smallest $\epsilon$ value for which bounds~\cite{xu2021fast} from the element-wise convex hull of the activations (with fixed intermediate bounds from \cite{Wong2018, zhang2018efficient}) fail to prove robustness. Both binary search procedures are run with a tolerance of $\epsilon_{\text{tol}}=0.1$. Denoting $\epsilon_{lb}$ as the smallest output from the two routines, and $\epsilon_{ub}$ as the largest, the following perturbation radius is chosen: $\epsilon = \frac{1}{3} \epsilon_{lb} + \frac{2}{3}\epsilon_{ub}$.

\paragraph*{Link} \url{https://github.com/stanleybak/vnncomp2021/tree/main/benchmarks/oval21}

\subsection{SRI ResNet A/B}

\paragraph*{Proposed by:} The MN-BaB team

\paragraph*{Motivation} While in previous instantiations of the VNN-COMP many benchmarks considered different architectures, thus allowing to judge the effect of architecture changes on the (relative) performance of different verifiers, none allowed for a direct comparison of the effect of different training methods. To enable such a comparison, we propose two benchmarks considering the same network architecture, trained using adversarial training of different strengths.

\paragraph*{Networks} We consider two residual ReLU networks~\cite{he2016deep} with 3 ResBlocks, preceded by one convolutional layer and followed by two fully-connected layers, yielding a total of eight ReLU layers. Both networks were trained using adversarial training, with the "A" version of the benchmark using a weaker attack compared to the "B" version.

\paragraph*{Specifications} We repeat the following process until we have collected 72 instances. We sample a random image from the CIFAR-10 test set, rejecting it immediately if it gets misclassified. If it gets classified correctly, we conduct a bisection search to find the smallest perturbation magnitude leading to misclassification under adversarial attacks (with a maximum of epsilon=0.005). We then generate a specification describing correct classification under an $\ell_\infty$-norm perturbation of at most $0.7$ times the previously found epsilon and allow a per sample timeout of 5 minutes.

\subsection{VGGNET16 2022}
\paragraph*{Proposed by} Stanley Bak, Stony Brook University

\paragraph*{Motivation} This benchmark tries to scale up the size of networks being analyzed by using the well-studied VGGNET-16 architecture~\cite{simonyan2014very} that runs on ImageNet. Input-output properties are proposed on pixel-level perturbations that can lead to image misclassification. 

\paragraph*{Networks} All properties are run on the same network, which includes 138 million parameters. The network features convolution layers, ReLU activation functions, as well as max pooling layers.

\paragraph*{Specifications} Properties analyzed ranged from single-pixel perturbations to perturbations on over 95k of the inputs. Full L-infinity perturbations were not used, as this caused an issue with the public specification file parser, which is a change we can address next year. A subset of the images was used to create the specifications, one from each category, which was randomly chosen to attack. Pixels to perturb were also randomly selected according to a random seed.

\paragraph*{Link} \url{https://github.com/stanleybak/vggnet16_benchmark2022/}


\subsection{MNIST FC}

\paragraph{Proposed by} The VeriNet team.

\paragraph*{Motivation} This benchmark contains fully connected networks with ReLU activation functions and varying depths. 

\paragraph*{Networks} The benchmark set consists of three fully-connected classification networks with 2, 4 and 6 
layers and 256 ReLU nodes in each layer trained on the MNIST dataset. The networks were first presented in a benchmark in VNN-COMP 2020~\cite{vnn-comp}.

\paragraph*{Specifications} We randomly sampled 15 correctly classified images from the MNIST test set. For each network and image, the specification was a correct classification under $l_\infty$ perturbations of at most $\epsilon = 0.03$ and $\epsilon = 0.05$. The timeouts were 2 minutes per instance for the 2-layer network and 5 minutes for the remaining two networks.

\subsection{Reach Prob Density}
\paragraph*{Proposed by} Stanley Bak, Stony Brook University

\paragraph*{Motivation} This benchmark analyzes neural networks used to learn how probability densities change through differential equations~\cite{meng2022learning}.
Given a initial uniform distribution in a box, for example, flows will tend to accumulate at certain states and become more sparse in other parts of the sapce.
The function computing these is learned using a neural network, which can then be used to efficiently query for probabilistic statements about the reachable states or to perform control~\cite{meng2022case}.

\paragraph*{Networks} All networks consist of fully connected layers and ReLU activation functions, with 4.6k to 70k parameters and 128-512 neurons, with a low input dimension (4-8 inputs). The three systems analyzed from the original work were the Vanderpol model, the Robot model, and the GCAS model.

\vbox{
\paragraph*{Specifications} Properties were created based on random ranges over the input and probabilities in the specification. For example, for the robot system, a random radius was chosen uniformly between 0.0 and 0.3, along with a random log probability between 0.05 and 0.3, and then a state was searched for near the obstacle at the origin with that radius above the probability threshold. More specifics on the other systems are given in the link below.

\paragraph*{Link} \url{https://github.com/stanleybak/reach_probability_benchmark2022/}
}

\subsection{RL Benchmarks}
\paragraph*{Proposed by} Veena Krish, Stony Brook University

\paragraph*{Motivation} This benchmark deals with open-loop sensor-noise robustness of reinforcement learning systems. Discrete output systems are examined. Given a specific state of the system, can bounded sensor noise cause the output command the change? These networks were chosen from work looking at closed-loop robustness where the open-loop problem would need to be repeatedly solved.

\paragraph*{Networks} All networks consist of fully connected layers and ReLU activation functions. All networks are small, with between 129 and 512 neurons.
Networks were created from the Dubin's rejoin task from the SafeRL benchmark suite~\cite{ravaioli2022safe}, as well as from the CartPole and LunarLander benchmarks from OpenAI gym~\cite{brockman2016openai}, trained using the StableBaselines3 library.

\paragraph*{Specifications} Specifications can be relatively complex disjunctions since the property is that any other command could have been chosen. The benchmarks check if the output command changes within some bounded noise.

\paragraph*{Link} \url{https://github.com/Ethos-lab/min-err-trajs-vnncomp-benchmarks}

\subsection{TLL Verify Bench}
\paragraph*{Proposed by} James Ferlez (University of California, Irvine)

\paragraph*{Motivation} This benchmark consists of Two-Level Lattice (TLL) NNs, which have been shown to be amenable to fast verification algorithms (e.g. \cite{FerlezKS22}). Thus, this benchmark was proposed as a means of comparing TLL-specific verification algorithms with general-purpose NN verification algorithms (i.e. algorithms that can verify arbitrary deep, fully-connected ReLU NNs).

\paragraph*{Networks}  The networks in this benchmark are a subset of the ones used in \cite[Experiment 3]{FerlezKS22}. Each of these TLL NNs has $n=2$ inputs and $m=1$ output. The architecture of a TLL NN is further specified by two parameters: $N$, the number of local linear functions, and $M$, the number of selector sets. This benchmark contains TLLs of sizes $N = M = 8, 16, 24, 32, 40, 48, 56, 64$, with $30$ randomly generated examples of each (the generation procedure is described in \cite[Section 6.1.1]{FerlezKS22}). At runtime, the specified verification timeout determines how many of these networks are included in the benchmark so as to achieve an overall 6-hour run time; this selection process is deterministic. Finally, a TLL NN has a natural representation using multiple computation paths \cite[Figure 1]{FerlezKS22}, but many tools are only compatible with fully-connected networks. Hence, the ONNX models in this benchmark implement TLL NNs by ``stacking'' these computation paths to make a fully connected NN (leading to sparse weight matrices: i.e. with many zero weights and biases). The \texttt{TLLnet} class (\url{https://github.com/jferlez/TLLnet}) contains the code necessary to generate these implementations via the \texttt{exportONNX} method.

\paragraph*{Specifications}  All specifications have as input constraints the hypercube $[-2,2]^2$. Since all networks have only a single output, the output properties consist of a randomly generated real number and a randomly generated inequality direction. Random output samples from the network are used to roughly ensure that the real number property has an equal likelihood of being within the output range of the NN and being outside of it (either above or below all NN outputs on the input constraint set). The inequality direction is generated independently and with each direction having an equal probability. This scheme biases the benchmark towards verification problems for which counterexamples exist. 

\paragraph*{Link} \url{https://github.com/jferlez/TLLVerifyBench}
\paragraph*{Commit}
199d2c26d0ec456e62906366b694a875a21ff7ef

\subsection{ACAS Xu (unscored)}
\paragraph{Networks} The ACASXu benchmark consists of ten properties defined over 45 neural networks used to issue turn advisories to aircraft to avoid collisions. The neural networks have 300 neurons arranged in 6 layers, with ReLU activation functions. There are five inputs corresponding to the aircraft states, and five network outputs, where the minimum output is used as the turn advisory the system ultimately produces.

\paragraph{Specifications} We use the original 10 properties~\cite{katz2017reluplex}, where properties 1-4 are checked on all 45 networks as was done in later work by the original authors~\cite{katz2019marabou}. Properties 5-10 are checked on a single network. The total number of benchmarks is therefore 186. The original verification times ranged from seconds to days---including some benchmark instances that did not finish. This year we used a timeout of around two minutes (116 seconds) for each property, in order to fit within a total maximum runtime of six hours.

\subsection{Cifar2020 (unscored)}

\paragraph*{Motivation} This benchmark combines two convolutional CIFAR10 networks from last year's VNN-COMP 2020 with a new, larger network with the goal to evaluate the progress made by the whole field of Neural Network verification.

\paragraph*{Networks} The two ReLU networks \texttt{cifar\_10\_2\_255} and \texttt{cifar\_10\_8\_255} with two convolutional and two fully-connected layers were trained for $\ell_\infty$ perturbations of $\epsilon = \frac{2}{255}$ and $\frac{8}{255}$, respectively, using COLT \cite{balunovic:20}  and the larger \texttt{ConvBig} with four convolutional and three fully-connected networks, was trained using adversarial training \cite{madry:17} and $\epsilon = \frac{2}{255}$.

\paragraph*{Specifications} We draw the first 100 images from the CIFAR10 test set and for every network reject incorrectly classified ones. For the remaining images, the specifications describe a correct classification under an $\ell_\infty$-norm perturbation of at most $\frac{2}{255}$ and $\frac{8}{255}$ for \texttt{cifar\_10\_2\_255} and \texttt{ConvBig} and \texttt{cifar\_10\_8\_255}, respectively and allow a per sample timeout of 5 minutes.

\newpage
\section{Results}
\label{sec:results}

Each tool was run on each of the benchmarks and produced a \texttt{csv} result file, that was provided as feedback to the tool authors using the online execution platform.
The final \texttt{csv} files for each tool as well as scoring scripts are available online: \url{https://github.com/ChristopherBrix/vnncomp2022_results}.
The results were analyzed automatically to compute scores and create the statistics presented in this section.

For purposes of scoring, recall that the minimum time after subtracting overhead was 1.0 seconds, so all times less than 1.0 get set to 1.0.
Then, the fastest tools are awarded bonus points as described in \Cref{sec:rules}.
Penalties usually occur when a tool produces an incorrect result, but in one case (with $\alpha,\!\beta$-CROWN) a penalty was awarded because no counterexample was provided by the competition version of the code (remedied in a later version).

\subsection{Overall Score}
\begin{figure}[h]
\centerline{\includegraphics[width=\textwidth]{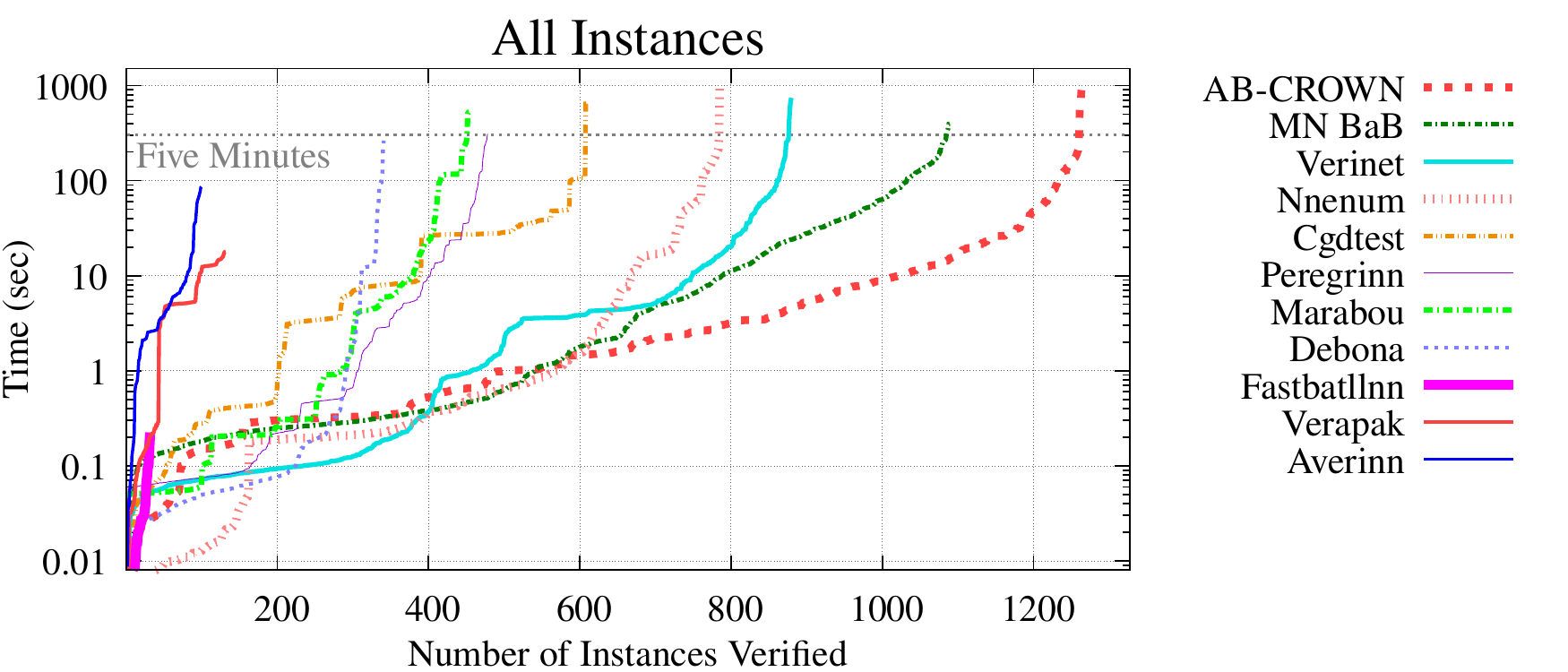}}
\caption{Cactus Plot for All Instances.}
\label{fig:all}
\end{figure}

\begin{wraptable}[116]{r}{0.28\textwidth}
\begin{center}
\vspace{-9mm}
\caption{Overall Score} \label{tab:score}
{\setlength{\tabcolsep}{2pt}
\vspace{-2mm}
\begin{tabular}{@{}lll@{}}
\toprule
\textbf{\# ~} & \textbf{Tool} & \textbf{Score}\\
\midrule
1 & $\alpha$,$\beta$-CROWN & 1274.9 \\
2 & \textsc{MN-BaB} & 1017.5 \\
3 & Verinet & 892.4 \\
4 & Nnenum & 534.0 \\
5 & Cgdtest & 408.4 \\
6 & Peregrinn & 399.0 \\
7 & Marabou & 372.2 \\
8 & Debona & 222.9 \\
9 & Fastbatllnn & 100.0 \\
10 & Verapak & 98.2 \\
11 & Averinn & 29.1 \\
\bottomrule
\end{tabular}
}
\end{center}
\end{wraptable}
The winner of the VNN-COMP 2022 is $\alpha,\!\beta$-CROWN, followed by \textsc{MN-BaB} and then Verinet. For a full ranking see \Cref{tab:score}. In \Cref{fig:all} we show the number of instances that were solved by the different verifiers within a certain runtime. For a more detailed summary of the results of the individual benchmarks see \Cref{sec:benchmark_results_scored} (scored) and \Cref{sec:benchmark_results_unscored} (unscored). For even more detailed per instance results see \Cref{sec:results_detailed}.

The most successful (top 3) tools all used GPU-based verifiers combining linear bound propagation with the branch-and-bound paradigm and participated in all benchmarks.
Similar to the previous year, several tools produced mismatching results (see \Cref{tab:incorrect_results}). However, thanks to the mandatory counterexample files, the ground truth could often be established allowing the identification of the incorrect verifiers. This highlighted in particular that CGDTest often incorrectly returned \texttt{UNSAT}, i.e., claimed a property holds when other tools were able to find counterexamples.


\clearpage
\subsection{Other Stats}

This section presents other statistics related to the measurements that are interesting but did not play a direct role in scoring this year.



\begin{table}[h]
  \begin{minipage}[t]{.4\linewidth}
    \begin{center}
    \caption{Overhead} \label{tab:overhead}
    {\setlength{\tabcolsep}{2pt}
    \begin{tabular}[h]{@{}llr@{}}
    \toprule
    \textbf{\# ~} & \textbf{Tool} & \textbf{Seconds}\\
    \midrule
    1 & Marabou & 0.2 \\
    2 & Fastbatllnn & 0.5 \\
    3 & Nnenum & 0.9 \\
    4 & Cgdtest & 1.3 \\
    5 & Peregrinn & 1.3 \\
    6 & Debona & 2.0 \\
    7 & Averinn & 3.1 \\
    8 & Verinet & 3.4 \\
    9 & Verapak & 4.6 \\
    10 & $\alpha$,$\beta$-CROWN & 6.7 \\
    11 & \textsc{MN-BaB} & 8.2 \\
    \bottomrule
    \end{tabular}
    }
    \end{center}
    \end{minipage}
\hfill
    \begin{minipage}[t]{0.5\linewidth}
        \begin{center}
        \caption{Num Benchmarks Participated} \label{tab:stats0}
        {\setlength{\tabcolsep}{2pt}
        \begin{tabular}[h]{@{}llr@{}}
        \toprule
        \textbf{\# ~} & \textbf{Tool} & \textbf{Count}\\
        \midrule
        1 & Verinet & 13 \\
        2 & \textsc{MN-BaB} & 13 \\
        3 & $\alpha$,$\beta$-CROWN & 13 \\
        4 & Cgdtest & 12 \\
        5 & Nnenum & 9 \\
        6 & Peregrinn & 7 \\
        7 & Marabou & 6 \\
        8 & Debona & 5 \\
        9 & Verapak & 3 \\
        10 & Fastbatllnn & 1 \\
        11 & Averinn & 1 \\
        \bottomrule
        \end{tabular}
        }
        \end{center}
    
    \end{minipage}
\end{table}

\begin{table}[h]
    \begin{minipage}[c]{0.35\textwidth}
        \begin{center}
        \caption{Num Instances Verified} \label{tab:stats1}
        {\setlength{\tabcolsep}{2pt}
        \begin{tabular}[h]{@{}llr@{}}
        \toprule
        \textbf{\# ~} & \textbf{Tool} & \textbf{Count}\\
        \midrule
        1 & $\alpha$,$\beta$-CROWN & 950 \\
        2 & \textsc{MN-BaB} & 812 \\
        3 & Verinet & 754 \\
        4 & Nnenum & 515 \\
        5 & Peregrinn & 478 \\
        6 & Marabou & 450 \\
        7 & Cgdtest & 405 \\
        8 & Debona & 341 \\
        9 & Verapak & 117 \\
        10 & Averinn & 100 \\
        11 & Fastbatllnn & 32 \\
        \bottomrule
        \end{tabular}
        }
        \end{center}
    \end{minipage}
    \hfill
    %
    \begin{minipage}[c]{0.25\linewidth}
    \begin{center}
    \caption{Num SAT} \label{tab:stats2}
    {\setlength{\tabcolsep}{2pt}
    \begin{tabular}[h]{@{}llr@{}}
    \toprule
    \textbf{\# ~} & \textbf{Tool} & \textbf{Count}\\
    \midrule
    1 & Verinet & 228 \\
    2 & \textsc{MN-BaB} & 227 \\
    3 & $\alpha$,$\beta$-CROWN & 227 \\
    4 & Nnenum & 196 \\
    5 & Peregrinn & 191 \\
    6 & Marabou & 148 \\
    7 & Debona & 138 \\
    8 & Cgdtest & 101 \\
    9 & Fastbatllnn & 21 \\
    10 & Averinn & 8 \\
    11 & Verapak & 6 \\
    \bottomrule
    \end{tabular}
    }
    \end{center}
    \end{minipage}
    \hfill
    \begin{minipage}{0.05\textwidth}    
    \end{minipage}
    \hfill
    \begin{minipage}[c]{0.25\linewidth}
    \begin{center}
    \caption{Num UNSAT} \label{tab:stats3}
    {\setlength{\tabcolsep}{2pt}
    \begin{tabular}[h]{@{}llr@{}}
    \toprule
    \textbf{\# ~} & \textbf{Tool} & \textbf{Count}\\
    \midrule
    1 & $\alpha$,$\beta$-CROWN & 723 \\
    2 & \textsc{MN-BaB} & 585 \\
    3 & Verinet & 526 \\
    4 & Nnenum & 319 \\
    5 & Cgdtest & 304 \\
    6 & Marabou & 302 \\
    7 & Peregrinn & 287 \\
    8 & Debona & 203 \\
    9 & Verapak & 111 \\
    10 & Averinn & 92 \\
    11 & Fastbatllnn & 11 \\
    \bottomrule
    \end{tabular}
    }
    \end{center}
    \end{minipage}
\end{table}


\begin{table}[h]
\begin{center}
\caption{Incorrect Results (or Missing CE ({*}))} \label{tab:incorrect_results}
{\setlength{\tabcolsep}{2pt}
\begin{threeparttable}[h]
\renewcommand{\TPTminimum}{\linewidth}
\begin{center}
\makebox[\linewidth]
{
\begin{tabular}[h]{@{}llr@{}}
\toprule
\textbf{\# ~} & \textbf{Tool} & \textbf{Count}\\
\midrule
1 & Cgdtest & 41 \\
2 & Verapak & 5 \\
3 & Marabou & 1 \\
4 & $\alpha$,$\beta$-CROWN\textsuperscript{*} & 1 \\
\bottomrule
\end{tabular}
}
\end{center}
\begin{tablenotes}
\centering
\item[*] Result was correct, but counterexample (CE) was not saved. See Table \ref{tab:benchmark_collins}.
\end{tablenotes}

\end{threeparttable}
}
\end{center}
\end{table}

\section{Conclusion and Ideas for Future Competitions}
\label{sec:conclusion}
This report summarizes the 3$^\text{rd}$ Verification of Neural Networks Competition (VNN-COMP), held in 2022.
While we observed a significant increase in the diversity, complexity, and scale of the proposed benchmarks, the best-performing tools seem to converge to GPU-enabled linear bound propagation methods using a branch-and-bound framework.
In addition to the standardization of input formats (\texttt{onnx} and \texttt{vnnlib}) and evaluation hardware, introduced for VNN-COMP 2021, VNN-COMP 2022 also standardized a format for counter-examples and introduced a fully automated evaluation pipeline, requiring authors to provide complete installation scripts.
We hope that this increased standardization and automatization does not only simplify the evaluation during the competition but also enables practitioners and researchers to more easily apply a range of state-of-the-art verification methods to their individual problems.

VNN-COMP 2022, successfully implemented a range of improvement opportunities identified during the previous iteration. These included requiring witnesses of found counter-examples to disambiguate tool disagreement, increasing automatization to enable a smoother final evaluation, and making a broader range of AWS instances available to allow for a better fit with tools' requirements. 
Further ideas for future competitions include the use of scored benchmarks specifically designed for year-on-year progress tracking, the reduction of tool tuning, a batch-processing mode, and more rigorous soundness evaluation.

\section*{Acknowledgements}
%

This competition is supported by a gift from the Lu Jin Family Foundation. 

This research was supported in part by the Air Force Research Laboratory Information Directorate, through the Air Force Office of Scientific Research Summer Faculty Fellowship
Program, Contract Numbers FA8750-15-3-6003, FA9550-15-0001 and FA9550-20-F-0005.
This material is based upon work supported by the Air Force Office of Scientific Research under award numbers FA9550-19-1-0288, FA9550-21-1-0121, and FA9550-22-1-0019, the National Science Foundation (NSF) under grant numbers 1918450, 1911017, 2028001, 2220401, and 2220426, and the Defense Advanced Research Projects Agency (DARPA) Assured Autonomy program through contract number FA8750-18-C-0089.
Any opinions, findings, and conclusions or recommendations expressed in this material are those of the author(s) and do not necessarily reflect the views of the United States Air Force, DARPA, nor NSF.

Tool authors listed in \Cref{sec:participants} participated in the preparation and review of this report.

\clearpage
\label{sect:bib}
\bibliographystyle{plain}
\bibliography{bib/nnv, bib/nnenum, bib/peregriNN, bib/verinet, bib/oval,bib/venus,bib/MIPVerify, bib/mnbab, bib/alpha-beta-CROWN,bib/debona,bib/dnnf,bib/nvjl,bib/nn4sys,bib/Marabou,bib/RPM,bib/AVeriNN,bib/VeRAPAk,bib/general}

\begin{thebibliography}{10}

\bibitem{bak2020vnn}
Stanley Bak.
\newblock Execution-guided overapproximation (ego) for improving scalability of
  neural network verification, 2020.

\bibitem{bak2021nnenum}
Stanley Bak.
\newblock nnenum: Verification of relu neural networks with optimized
  abstraction refinement.
\newblock In {\em NASA Formal Methods Symposium}, pages 19--36. Springer, 2021.

\bibitem{bak2020cav}
Stanley Bak, Hoang-Dung Tran, Kerianne Hobbs, and Taylor~T. Johnson.
\newblock Improved geometric path enumeration for verifying {ReLU} neural
  networks.
\newblock In {\em 32nd International Conference on Computer-Aided Verification
  (CAV)}, July 2020.

\bibitem{brix2020debona}
Christopher Brix and Thomas Noll.
\newblock Debona: Decoupled boundary network analysis for tighter bounds and
  faster adversarial robustness proofs.
\newblock {\em CoRR}, abs/2006.09040, 2020.

\bibitem{brockman2016openai}
Greg Brockman, Vicki Cheung, Ludwig Pettersson, Jonas Schneider, John Schulman,
  Jie Tang, and Wojciech Zaremba.
\newblock Openai gym.
\newblock {\em arXiv preprint arXiv:1606.01540}, 2016.

\bibitem{bunel2020lagrangian}
Rudy Bunel, Alessandro De~Palma, Alban Desmaison, Krishnamurthy Dvijotham,
  Pushmeet Kohli, Philip~HS Torr, and M~Pawan Kumar.
\newblock Lagrangian decomposition for neural network verification.
\newblock {\em Conference on Uncertainty in Artificial Intelligence}, 2020.

\bibitem{bunelunified2018}
Rudy Bunel, Ilker Turkaslan, Philip~HS Torr, Pushmeet Kohli, and M~Pawan Kumar.
\newblock A unified view of piecewise linear neural network verification.
\newblock {\em Advances in Neural Information Processing Systems}, 2018.

\bibitem{depalma2021scaling}
Alessandro De~Palma, Harkirat~Singh Behl, Rudy Bunel, Philip H.~S. Torr, and
  M.~Pawan Kumar.
\newblock Scaling the convex barrier with active sets.
\newblock In {\em International Conference on Learning Representations}, 2021.

\bibitem{depalma2021sparsealgos}
Alessandro De~Palma, Harkirat~Singh Behl, Rudy Bunel, Philip H.~S. Torr, and
  M.~Pawan Kumar.
\newblock Scaling the convex barrier with sparse dual algorithms.
\newblock {\em arXiv preprint arXiv:2101.05844}, 2021.

\bibitem{depalma2021improved}
Alessandro De~Palma, Rudy Bunel, Alban Desmaison, Krishnamurthy Dvijotham,
  Pushmeet Kohli, Philip~HS Torr, and M~Pawan Kumar.
\newblock Improved branch and bound for neural network verification via
  lagrangian decomposition.
\newblock {\em arXiv preprint arXiv:2104.06718}, 2021.

\bibitem{dong2018boosting}
Yinpeng Dong, Fangzhou Liao, Tianyu Pang, Hang Su, Jun Zhu, Xiaolin Hu, and
  Jianguo Li.
\newblock Boosting adversarial attacks with momentum.
\newblock In {\em Proceedings of the IEEE conference on computer vision and
  pattern recognition}, pages 9185--9193, 2018.

\bibitem{FerlezKS22}
James Ferlez, Haitham Khedr, and Yasser Shoukry.
\newblock Fast {BATLLNN:} fast box analysis of two-level lattice neural
  networks.
\newblock In Ezio Bartocci and Sylvie Putot, editors, {\em {HSCC} '22: 25th
  {ACM} International Conference on Hybrid Systems: Computation and Control,
  Milan, Italy, May 4 - 6, 2022}, pages 23:1--23:11. {ACM}, 2022.

\bibitem{mnbab}
Claudio Ferrari, Mark~Niklas M{\"{u}}ller, Nikola Jovanovic, and Martin~T.
  Vechev.
\newblock Complete verification via multi-neuron relaxation guided
  branch-and-bound.
\newblock In {\em The Tenth International Conference on Learning
  Representations, {ICLR} 2022, Virtual Event, April 25-29, 2022}.
  OpenReview.net, 2022.

\bibitem{gurobi}
{Gurobi Optimization, LLC}.
\newblock {Gurobi Optimizer Reference Manual}, 2022.

\bibitem{he2022Characterizing}
Haoyu He, Tianhao Wei, Huan Zhang, Changliu Liu, and Cheng Tan.
\newblock Characterizing neural network verification for systems with
  {NN4SYSBench}.
\newblock {\em 1st Workshop on Formal Verification of Machine Learning}, 2022.

\bibitem{he2016deep}
Kaiming He, Xiangyu Zhang, Shaoqing Ren, and Jian Sun.
\newblock Deep residual learning for image recognition.
\newblock In {\em Proceedings of the IEEE conference on computer vision and
  pattern recognition}, pages 770--778, 2016.

\bibitem{Henriksen+21}
P.~Henriksen, K.~Hammernik, D.~Rueckert, and A.~Lomuscio.
\newblock Bias field robustness verification of large neural image classifiers.
\newblock In {\em Proceedings of the 32nd British Machine Vision Conference
  ({BMVC}21)}. {BMVA} Press, 2021.

\bibitem{HenriksenLomuscio20}
P.~Henriksen and A.~Lomuscio.
\newblock Efficient neural network verification via adaptive refinement and
  adversarial search.
\newblock In {\em Proceedings of the 24th European Conference on Artificial
  Intelligence (ECAI20)}, 2020.

\bibitem{HenriksenLomuscio21}
P.~Henriksen and A.~Lomuscio.
\newblock Deepsplit: An efficient splitting method for neural network
  verification via indirect effect analysis.
\newblock In {\em Proceedings of the 30th International Joint Conference on
  Artificial Intelligence (IJCAI21)}, To appear, August 2021.

\bibitem{jaeckle2021generating}
Florian Jaeckle and M~Pawan Kumar.
\newblock Generating adversarial examples with graph neural networks.
\newblock {\em Conference on Uncertainty in Artificial Intelligence}, 2021.

\bibitem{jaeckle2021neural}
Florian Jaeckle, Jingyue Lu, and M~Pawan Kumar.
\newblock Neural network branch-and-bound for neural network verification.
\newblock {\em arXiv preprint arXiv:2107.12855}, 2021.

\bibitem{katz2017reluplex}
Guy Katz, Clark Barrett, David~L Dill, Kyle Julian, and Mykel~J Kochenderfer.
\newblock Reluplex: An efficient smt solver for verifying deep neural networks.
\newblock In {\em International Conference on Computer Aided Verification},
  pages 97--117. Springer, 2017.

\bibitem{KatzHIJLLSTWZDK19}
Guy Katz, Derek~A. Huang, Duligur Ibeling, Kyle Julian, Christopher Lazarus,
  Rachel Lim, Parth Shah, Shantanu Thakoor, Haoze Wu, Aleksandar Zeljic,
  David~L. Dill, Mykel~J. Kochenderfer, and Clark~W. Barrett.
\newblock The marabou framework for verification and analysis of deep neural
  networks.
\newblock In Isil Dillig and Serdar Tasiran, editors, {\em Computer Aided
  Verification - 31st International Conference, {CAV} 2019, New York City, NY,
  USA, July 15-18, 2019, Proceedings, Part {I}}, volume 11561 of {\em Lecture
  Notes in Computer Science}, pages 443--452. Springer, 2019.

\bibitem{katz2019marabou}
Guy Katz, Derek~A Huang, Duligur Ibeling, Kyle Julian, Christopher Lazarus,
  Rachel Lim, Parth Shah, Shantanu Thakoor, Haoze Wu, Aleksandar Zelji{\'c},
  et~al.
\newblock The marabou framework for verification and analysis of deep neural
  networks.
\newblock In {\em International Conference on Computer Aided Verification},
  pages 443--452. Springer, 2019.

\bibitem{khedr2021peregrinn}
Haitham Khedr, James Ferlez, and Yasser Shoukry.
\newblock Peregrinn: Penalized-relaxation greedy neural network verifier.
\newblock In {\em International Conference on Computer Aided Verification},
  pages 287--300. Springer, 2021.

\bibitem{kraska18case}
Tim Kraska, Alex Beutel, Ed~H Chi, Jeffrey Dean, and Neoklis Polyzotis.
\newblock The case for learned index structures.
\newblock In {\em Proceedings of the 2018 International Conference on
  Management of Data}, 2018.

\bibitem{Lu2020Neural}
Jingyue Lu and M~Pawan Kumar.
\newblock Neural network branching for neural network verification.
\newblock In {\em International Conference on Learning Representations}, 2020.

\bibitem{madry2017towards}
Aleksander Madry, Aleksandar Makelov, Ludwig Schmidt, Dimitris Tsipras, and
  Adrian Vladu.
\newblock Towards deep learning models resistant to adversarial attacks.
\newblock {\em arXiv preprint arXiv:1706.06083}, 2017.

\bibitem{madry:17}
Aleksander Madry, Aleksandar Makelov, Ludwig Schmidt, Dimitris Tsipras, and
  Adrian Vladu.
\newblock Towards deep learning models resistant to adversarial attacks.
\newblock In {\em Proc. International Conference on Learning Representations
  (ICLR)}, 2018.

\bibitem{meng2022case}
Yue Meng, Zeng Qiu, Md~Tawhid~Bin Waez, and Chuchu Fan.
\newblock Case studies for computing density of reachable states for safe
  autonomous motion planning.
\newblock In {\em NASA Formal Methods Symposium}, pages 251--271. Springer,
  2022.

\bibitem{meng2022learning}
Yue Meng, Dawei Sun, Zeng Qiu, Md~Tawhid~Bin Waez, and Chuchu Fan.
\newblock Learning density distribution of reachable states for autonomous
  systems.
\newblock In {\em Conference on Robot Learning}, pages 124--136. PMLR, 2022.

\bibitem{balunovic:20}
Martin~Vechev Mislav~Balunovic.
\newblock Adversarial training and provable defenses: Bridging the gap.
\newblock In {\em Proc. International Conference on Learning Representations
  (ICLR)}, 2020.

\bibitem{mueller2021prima}
Mark~Niklas M{\"u}ller, Gleb Makarchuk, Gagandeep Singh, Markus P{\"u}schel,
  and Martin Vechev.
\newblock Prima: Precise and general neural network certification via
  multi-neuron convex relaxations.
\newblock {\em arXiv preprint arXiv:2103.03638}, 2021.

\bibitem{PaszkeGMLBCKLGA19}
Adam Paszke, Sam Gross, Francisco Massa, Adam Lerer, James Bradbury, Gregory
  Chanan, Trevor Killeen, Zeming Lin, Natalia Gimelshein, Luca Antiga, Alban
  Desmaison, Andreas K{\"{o}}pf, Edward~Z. Yang, Zachary DeVito, Martin Raison,
  Alykhan Tejani, Sasank Chilamkurthy, Benoit Steiner, Lu~Fang, Junjie Bai, and
  Soumith Chintala.
\newblock Pytorch: An imperative style, high-performance deep learning library.
\newblock In Hanna~M. Wallach, Hugo Larochelle, Alina Beygelzimer, Florence
  d'Alch{\'{e}}{-}Buc, Emily~B. Fox, and Roman Garnett, editors, {\em Advances
  in Neural Information Processing Systems 32: Annual Conference on Neural
  Information Processing Systems 2019, NeurIPS 2019, December 8-14, 2019,
  Vancouver, BC, Canada}, pages 8024--8035, 2019.

\bibitem{prabhakar2019abstraction}
Pavithra Prabhakar and Zahra Rahimi~Afzal.
\newblock Abstraction based output range analysis for neural networks.
\newblock {\em Advances in Neural Information Processing Systems}, 32, 2019.

\bibitem{ravaioli2022safe}
Umberto~J Ravaioli, James Cunningham, John McCarroll, Vardaan Gangal, Kyle
  Dunlap, and Kerianne~L Hobbs.
\newblock Safe reinforcement learning benchmark environments for aerospace
  control systems.
\newblock In {\em 2022 IEEE Aerospace Conference (AERO)}, pages 1--20. IEEE,
  2022.

\bibitem{ronneberger2015u}
Olaf Ronneberger, Philipp Fischer, and Thomas Brox.
\newblock U-net: Convolutional networks for biomedical image segmentation.
\newblock In {\em International Conference on Medical image computing and
  computer-assisted intervention}, pages 234--241. Springer, 2015.

\bibitem{shriver-etal:CAV:2021:dnnv}
David Shriver, Sebastian~G. Elbaum, and Matthew~B. Dwyer.
\newblock {DNNV:} {A} framework for deep neural network verification.
\newblock In Alexandra Silva and K.~Rustan~M. Leino, editors, {\em Computer
  Aided Verification - 33rd International Conference, {CAV} 2021, Virtual
  Event, July 20-23, 2021, Proceedings, Part {I}}, volume 12759 of {\em Lecture
  Notes in Computer Science}, pages 137--150. Springer, 2021.

\bibitem{simonyan2014very}
Karen Simonyan and Andrew Zisserman.
\newblock Very deep convolutional networks for large-scale image recognition.
\newblock {\em arXiv preprint arXiv:1409.1556}, 2014.

\bibitem{singh2019krelu}
Gagandeep Singh, Rupanshu Ganvir, Markus P\"{u}schel, and Martin Vechev.
\newblock Beyond the single neuron convex barrier for neural network
  certification.
\newblock In {\em Advances in Neural Information Processing Systems 32}, pages
  15098--15109. Curran Associates, Inc., 2019.

\bibitem{DeepPoly:19}
Gagandeep Singh, Timon Gehr, Markus P{\"{u}}schel, and Martin~T. Vechev.
\newblock An abstract domain for certifying neural networks.
\newblock {\em Proc. {ACM} Program. Lang.}, 3({POPL}):41:1--41:30, 2019.

\bibitem{singh2019refinement}
Gagandeep Singh, Timon Gehr, Markus P{\"{u}}schel, and Martin~T. Vechev.
\newblock Boosting robustness certification of neural networks.
\newblock In {\em 7th International Conference on Learning Representations,
  {ICLR} 2019, New Orleans, LA, USA, May 6-9, 2019}. OpenReview.net, 2019.

\bibitem{Smith2021}
Joshua Smith, Jarom Allan, Viswanathan Swaminathan, and Zhen Zhang.
\newblock Refutation-based adversarial robustness verification of deep neural
  networks.
\newblock In {\em Formal Methods for ML-Enabled Autonomous Systems}, 7 2021.

\bibitem{Tjeng2019EvaluatingRO}
Vincent Tjeng, Kai~Y. Xiao, and Russ Tedrake.
\newblock Evaluating robustness of neural networks with mixed integer
  programming.
\newblock In {\em ICLR}, 2019.

\bibitem{TjengXT19}
Vincent Tjeng, Kai~Yuanqing Xiao, and Russ Tedrake.
\newblock Evaluating robustness of neural networks with mixed integer
  programming.
\newblock In {\em 7th International Conference on Learning Representations,
  {ICLR} 2019, New Orleans, LA, USA, May 6-9, 2019}. OpenReview.net, 2019.

\bibitem{tran2020cav}
Hoang-Dung Tran, Stanley Bak, Weiming Xiang, and Taylor~T. Johnson.
\newblock Verification of deep convolutional neural networks using imagestars.
\newblock In {\em 32nd International Conference on Computer-Aided Verification
  (CAV)}. Springer, July 2020.

\bibitem{tran2019fm}
Hoang-Dung Tran, Patrick Musau, Diego~Manzanas Lopez, Xiaodong Yang, Luan~Viet
  Nguyen, Weiming Xiang, and Taylor~T. Johnson.
\newblock Star-based reachability analysis for deep neural networks.
\newblock In {\em 23rd International Symposium on Formal Methods (FM'19)}.
  Springer International Publishing, October 2019.

\bibitem{tran2021robustness}
Hoang-Dung Tran, Neelanjana Pal, Patrick Musau, Diego~Manzanas Lopez, Nathaniel
  Hamilton, Xiaodong Yang, Stanley Bak, and Taylor~T Johnson.
\newblock Robustness verification of semantic segmentation neural networks
  using relaxed reachability.
\newblock In {\em International Conference on Computer Aided Verification},
  pages 263--286. Springer, 2021.

\bibitem{vnn-comp}
VNN-COMP.
\newblock International verification of neural networks competition
  ({VNN-COMP}).
\newblock {\em Verification of Neural Networks workshop at the International
  Conference on Computer-Aided Verification}, 2020.

\bibitem{wang2021betacrown}
Shiqi Wang, Huan Zhang, Kaidi Xu, Xue Lin, Suman Jana, Cho-Jui Hsieh, and Zico
  Kolter.
\newblock {Beta-CROWN}: Efficient bound propagation with per-neuron split
  constraints for complete and incomplete neural network verification.
\newblock {\em arXiv preprint arXiv:2103.06624}, 2021.

\bibitem{Wong2018}
Eric Wong and Zico Kolter.
\newblock Provable defenses against adversarial examples via the convex outer
  adversarial polytope.
\newblock {\em ICML}, 2018.

\bibitem{vegas}
Haoze Wu, Clark Barrett, Mahmood Sharif, Nina Narodytska, and Gagandeep Singh.
\newblock Scalable verification of gnn-based job schedulers.
\newblock 6(OOPSLA2), oct 2022.

\bibitem{wu2020parallelization}
Haoze Wu, Alex Ozdemir, Aleksandar Zeljic, Kyle Julian, Ahmed Irfan, Divya
  Gopinath, Sadjad Fouladi, Guy Katz, Corina Pasareanu, and Clark Barrett.
\newblock Parallelization techniques for verifying neural networks.
\newblock In {\em \# PLACEHOLDER\_PARENT\_METADATA\_VALUE\#}, volume~1, pages
  128--137. TU Wien Academic Press, 2020.

\bibitem{wu2022toward}
Haoze Wu, Teruhiro Tagomori, Alexander Robey, Fengjun Yang, Nikolai Matni,
  George Pappas, Hamed Hassani, Corina Pasareanu, and Clark Barrett.
\newblock Toward certified robustness against real-world distribution shifts.
\newblock {\em arXiv preprint arXiv:2206.03669}, 2022.

\bibitem{wu2022efficient}
Haoze Wu, Aleksandar Zelji{\'c}, Guy Katz, and Clark Barrett.
\newblock Efficient neural network analysis with sum-of-infeasibilities.
\newblock In {\em International Conference on Tools and Algorithms for the
  Construction and Analysis of Systems}, pages 143--163. Springer, 2022.

\bibitem{xu2020automatic}
Kaidi Xu, Zhouxing Shi, Huan Zhang, Yihan Wang, Kai-Wei Chang, Minlie Huang,
  Bhavya Kailkhura, Xue Lin, and Cho-Jui Hsieh.
\newblock Automatic perturbation analysis for scalable certified robustness and
  beyond.
\newblock {\em Advances in Neural Information Processing Systems}, 33, 2020.

\bibitem{xu2021fast}
Kaidi Xu, Huan Zhang, Shiqi Wang, Yihan Wang, Suman Jana, Xue Lin, and Cho-Jui
  Hsieh.
\newblock {Fast and Complete}: Enabling complete neural network verification
  with rapid and massively parallel incomplete verifiers.
\newblock In {\em International Conference on Learning Representations}, 2021.

\bibitem{zhang2019towards}
Huan Zhang, Hongge Chen, Chaowei Xiao, Sven Gowal, Robert Stanforth, Bo~Li,
  Duane Boning, and Cho-Jui Hsieh.
\newblock Towards stable and efficient training of verifiably robust neural
  networks.
\newblock {\em arXiv preprint arXiv:1906.06316}, 2019.

\bibitem{zhang2022general}
Huan Zhang*, Shiqi Wang*, Kaidi Xu*, Linyi Li, Bo~Li, Suman Jana, Cho-Jui
  Hsieh, and J~Zico Kolter.
\newblock General cutting planes for bound-propagation-based neural network
  verification.
\newblock {\em Advances in Neural Information Processing Systems (NeurIPS)},
  2022.

\bibitem{zhang2018efficient}
Huan Zhang, Tsui-Wei Weng, Pin-Yu Chen, Cho-Jui Hsieh, and Luca Daniel.
\newblock Efficient neural network robustness certification with general
  activation functions.
\newblock {\em Advances in Neural Information Processing Systems},
  31:4939--4948, 2018.

\end{thebibliography}


\appendix

\clearpage
\section{Extended Results}
In this section, we provide more fine-grained results.

\subsection{Scored Benchmarks}
\label{sec:benchmark_results_scored}


\begin{table}[h]
\begin{center}
\caption{Benchmark \texttt{carvana-unet-2022}} 
{\setlength{\tabcolsep}{2pt}
\begin{tabular}[h]{@{}llllllrr@{}}
\toprule
\textbf{\# ~} & \textbf{Tool} & \textbf{Verified} & \textbf{Falsified} & \textbf{Fastest} & \textbf{Penalty} & \textbf{Score} & \textbf{Percent}\\
\midrule
1 & $\alpha$,$\beta$ Crown & 39 & 0 & 39 & 0 & 468 & 100.0\% \\
2 & \textsc{MN-BaB} & 19 & 0 & 0 & 0 & 209 & 44.7\% \\
3 & Verinet & 3 & 0 & 0 & 0 & 30 & 6.4\% \\
\bottomrule
\end{tabular}
}
\end{center}
\end{table}

\begin{figure}[h]
\centerline{\includegraphics[width=\textwidth]{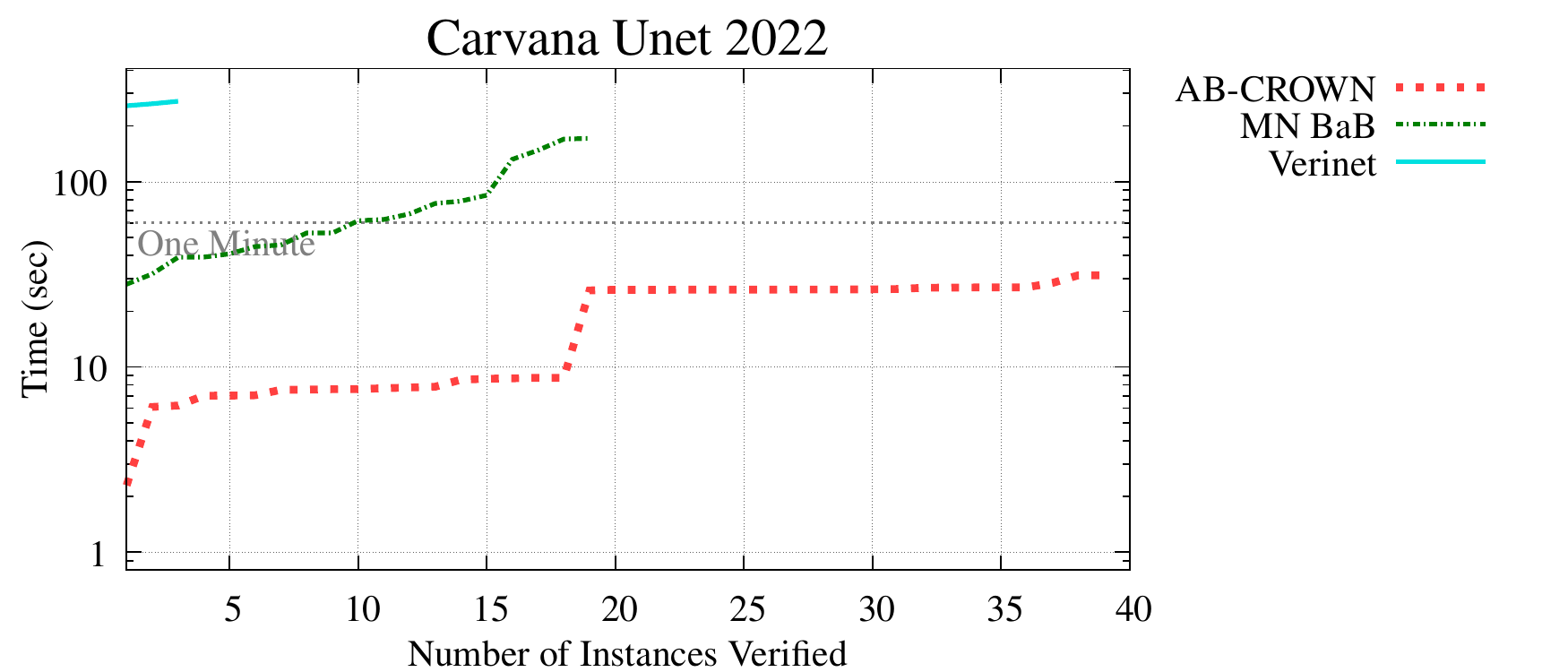}}
\caption{Cactus Plot for Carvana Unet 2022.}
\end{figure}


\begin{table}[h]
\begin{center}
\caption{Benchmark \texttt{cifar100-tinyimagenet-resnet}} 
{\setlength{\tabcolsep}{2pt}
\begin{tabular}[h]{@{}llllllrr@{}}
\toprule
\textbf{\# ~} & \textbf{Tool} & \textbf{Verified} & \textbf{Falsified} & \textbf{Fastest} & \textbf{Penalty} & \textbf{Score} & \textbf{Percent}\\
\midrule
1 & $\alpha$,$\beta$ Crown & 69 & 0 & 56 & 0 & 813 & 100.0\% \\
2 & Cgdtest & 95 & 0 & 28 & 3 & 725 & 89.2\% \\
3 & \textsc{MN-BaB} & 60 & 3 & 10 & 0 & 674 & 82.9\% \\
4 & Verinet & 48 & 3 & 6 & 0 & 540 & 66.4\% \\
\bottomrule
\end{tabular}
}
\end{center}
\end{table}

\begin{figure}[h]
\centerline{\includegraphics[width=\textwidth]{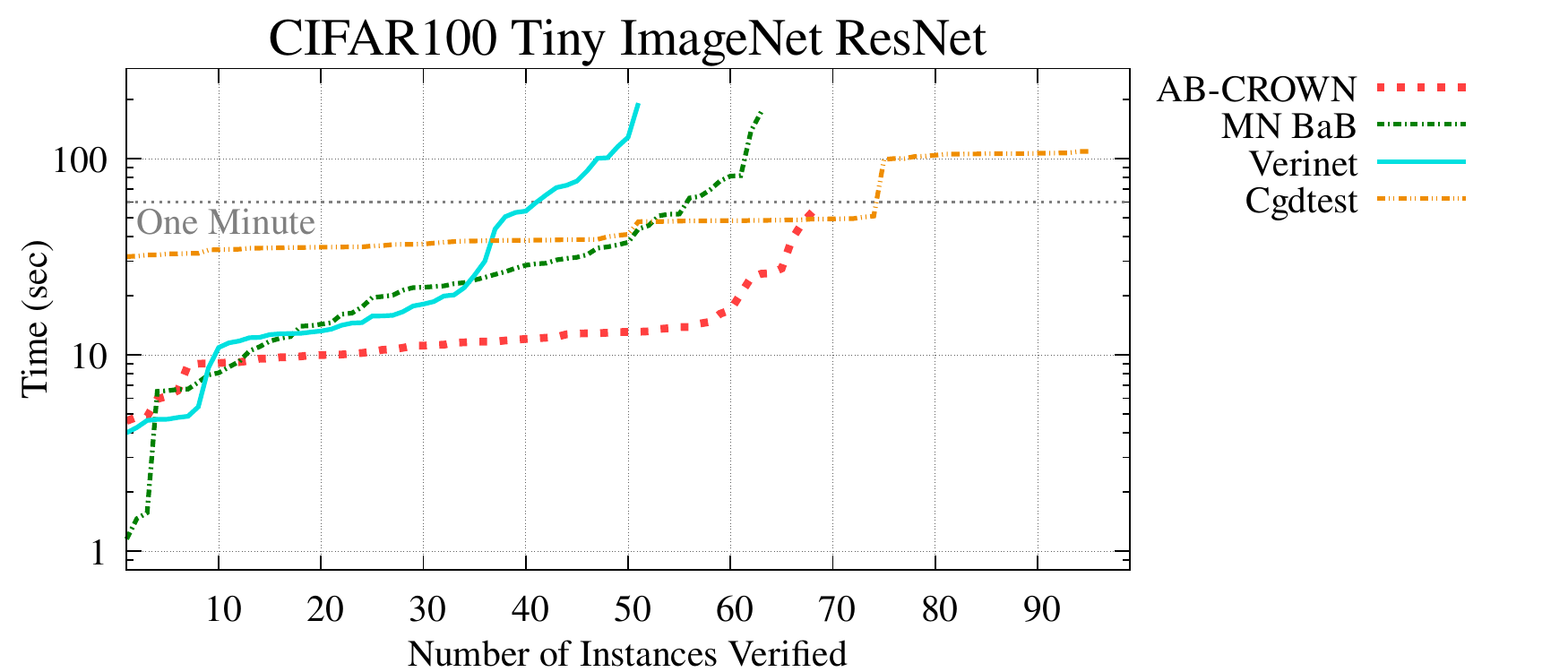}}
\caption{Cactus Plot for CIFAR100 Tiny ImageNet ResNet.}
\end{figure}


\begin{table}[h]
\begin{center}
\caption{Benchmark \texttt{cifar-biasfield}} 
{\setlength{\tabcolsep}{2pt}
\begin{tabular}[h]{@{}llllllrr@{}}
\toprule
\textbf{\# ~} & \textbf{Tool} & \textbf{Verified} & \textbf{Falsified} & \textbf{Fastest} & \textbf{Penalty} & \textbf{Score} & \textbf{Percent}\\
\midrule
1 & $\alpha$,$\beta$ Crown & 69 & 1 & 1 & 0 & 736 & 100.0\% \\
2 & Cgdtest & 71 & 0 & 55 & 1 & 731 & 99.3\% \\
3 & Verinet & 69 & 1 & 0 & 0 & 721 & 98.0\% \\
4 & Verapak & 71 & 0 & 0 & 1 & 635 & 86.3\% \\
5 & \textsc{MN-BaB} & 36 & 1 & 17 & 0 & 404 & 54.9\% \\
6 & Marabou & 27 & 0 & 0 & 0 & 270 & 36.7\% \\
7 & Nnenum & 4 & 0 & 0 & 0 & 43 & 5.8\% \\
\bottomrule
\end{tabular}
}
\end{center}
\end{table}

\begin{figure}[h]
\centerline{\includegraphics[width=\textwidth]{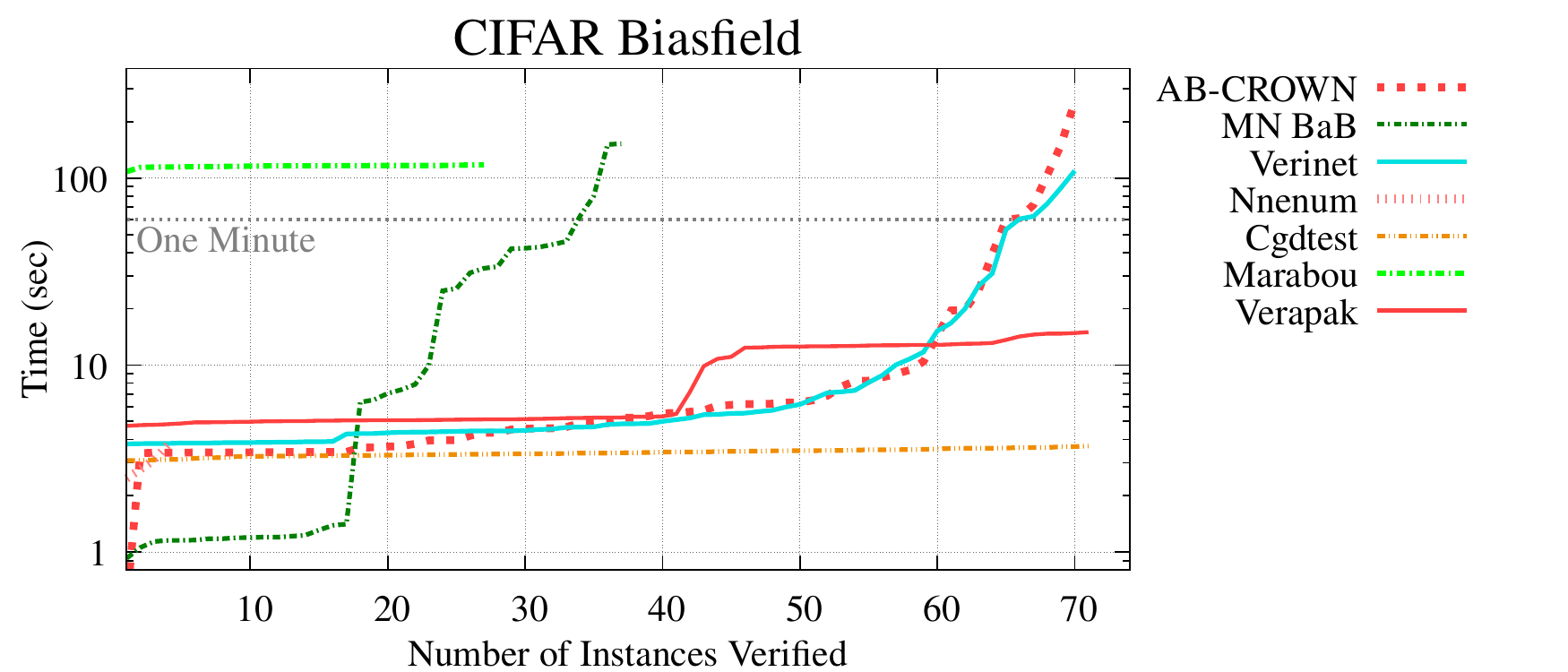}}
\caption{Cactus Plot for CIFAR Biasfield.}
\end{figure}


\begin{table}[h]
\begin{center}
\caption{Benchmark \texttt{collins-rul-cnn}} \label{tab:benchmark_collins}
{\setlength{\tabcolsep}{2pt}
\begin{threeparttable}[h]
\renewcommand{\TPTminimum}{\linewidth}
\begin{center}
\makebox[\linewidth]
{
\begin{tabular}[h]{@{}llllllrr@{}}
\toprule
\textbf{\# ~} & \textbf{Tool} & \textbf{Verified} & \textbf{Falsified} & \textbf{Fastest} & \textbf{Penalty} & \textbf{Score} & \textbf{Percent}\\
\midrule
1 & Nnenum & 16 & 45 & 58 & 0 & 727 & 100.0\% \\
2 & \textsc{MN-BaB} & 16 & 44 & 57 & 0 & 715 & 98.3\% \\
3 & $\alpha$,$\beta$ Crown & 15 & 45 & 56 & 1\textsuperscript{*} & 612 & 84.2\% \\
4 & Verinet & 16 & 43 & 0 & 0 & 590 & 81.2\% \\
5 & Peregrinn & 14 & 42 & 0 & 0 & 560 & 77.0\% \\
6 & Cgdtest & 1 & 42 & 43 & 15 & -984 & 0\% \\
\bottomrule
\end{tabular}
}
\end{center}
\begin{tablenotes}
\item[*] During the competition, Cgdtest reported UNSAT for the 20th property, conflicting with SAT reported by $\alpha,\!\beta$-CROWN. $\alpha,\!\beta$-CROWN team provided a counterexample (CE) after scoring, proving the ground truth to be SAT. However, $\alpha,\!\beta$-CROWN was penalized because the CE was not saved to disk during the competition.
\end{tablenotes}
\end{threeparttable}
}
\end{center}
\end{table}

\begin{figure}[h]
\centerline{\includegraphics[width=\textwidth]{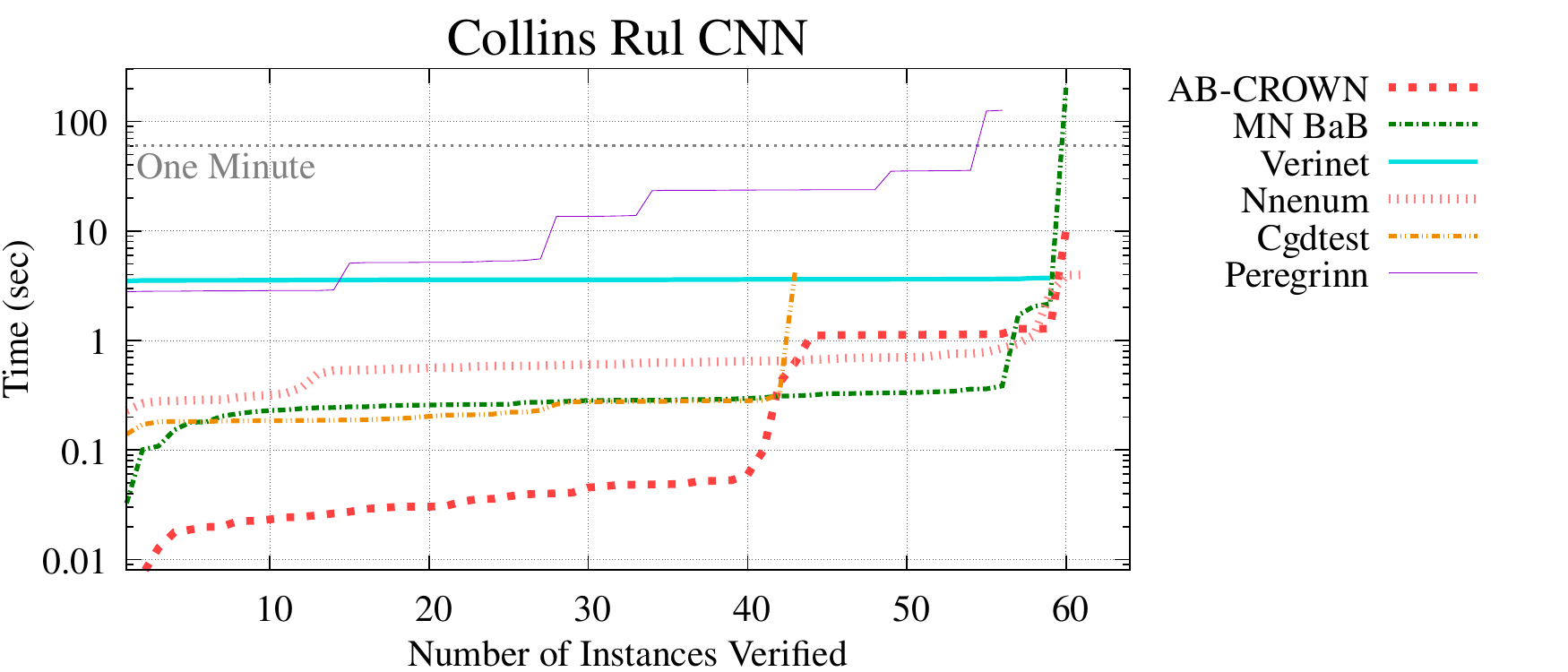}}
\caption{Cactus Plot for Collins Rul CNN.}
\end{figure}


\begin{table}[h]
\begin{center}
\caption{Benchmark \texttt{mnist-fc}} 
{\setlength{\tabcolsep}{2pt}
\begin{tabular}[h]{@{}llllllrr@{}}
\toprule
\textbf{\# ~} & \textbf{Tool} & \textbf{Verified} & \textbf{Falsified} & \textbf{Fastest} & \textbf{Penalty} & \textbf{Score} & \textbf{Percent}\\
\midrule
1 & $\alpha$,$\beta$ Crown & 66 & 18 & 53 & 0 & 963 & 100.0\% \\
2 & Verinet & 53 & 18 & 50 & 0 & 817 & 84.8\% \\
3 & \textsc{MN-BaB} & 53 & 18 & 47 & 0 & 804 & 83.5\% \\
4 & Debona & 48 & 18 & 38 & 0 & 737 & 76.5\% \\
5 & Nnenum & 48 & 11 & 29 & 0 & 649 & 67.4\% \\
6 & Marabou & 44 & 16 & 0 & 0 & 600 & 62.3\% \\
7 & Peregrinn & 27 & 11 & 7 & 0 & 394 & 40.9\% \\
8 & Cgdtest & 66 & 3 & 23 & 5 & 241 & 25.0\% \\
9 & Verapak & 40 & 2 & 42 & 4 & 104 & 10.8\% \\
\bottomrule
\end{tabular}
}
\end{center}
\end{table}

\begin{figure}[h]
\centerline{\includegraphics[width=\textwidth]{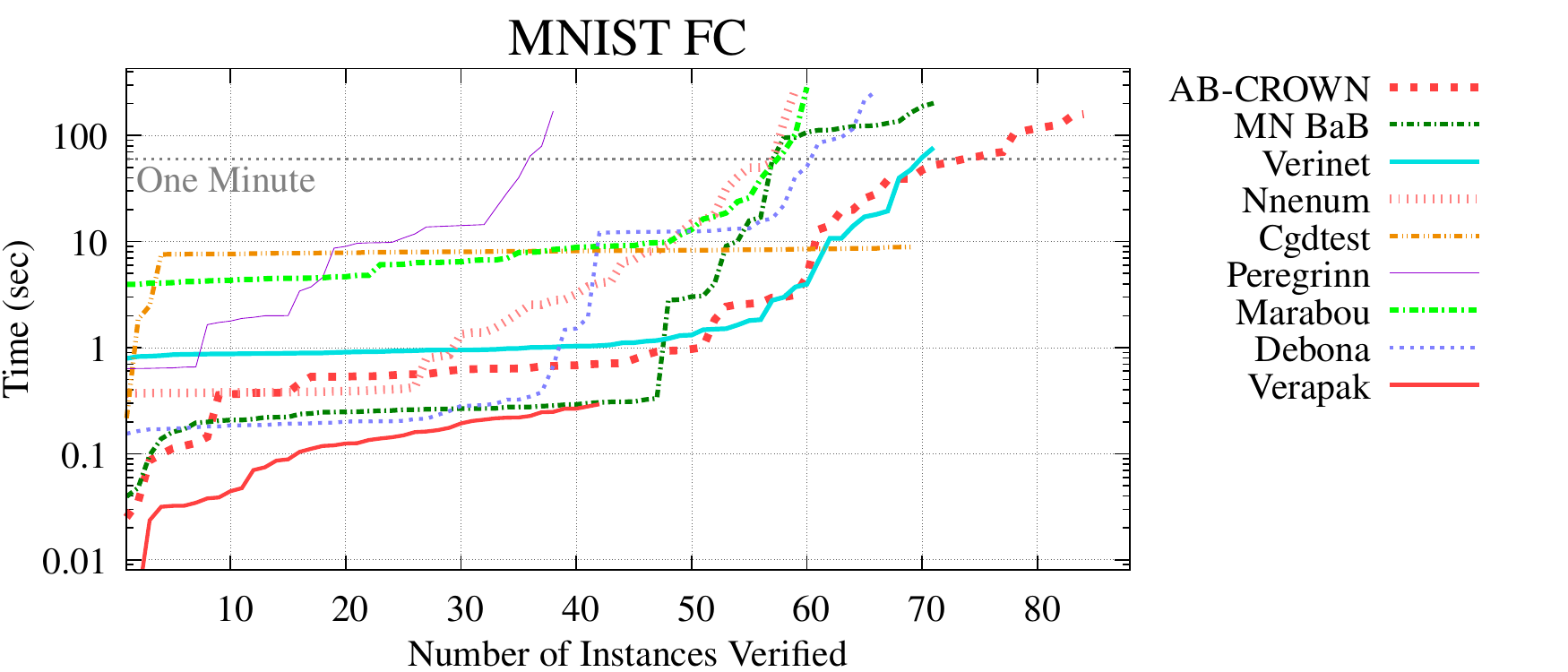}}
\caption{Cactus Plot for MNIST FC.}
\end{figure}


\begin{table}[h]
\begin{center}
\caption{Benchmark \texttt{nn4sys}} 
{\setlength{\tabcolsep}{2pt}
\begin{tabular}[h]{@{}llllllrr@{}}
\toprule
\textbf{\# ~} & \textbf{Tool} & \textbf{Verified} & \textbf{Falsified} & \textbf{Fastest} & \textbf{Penalty} & \textbf{Score} & \textbf{Percent}\\
\midrule
1 & $\alpha$,$\beta$ Crown & 152 & 0 & 132 & 0 & 1799 & 100.0\% \\
2 & \textsc{MN-BaB} & 106 & 0 & 8 & 0 & 1140 & 63.4\% \\
3 & Verinet & 57 & 0 & 43 & 0 & 661 & 36.7\% \\
4 & Peregrinn & 24 & 0 & 22 & 0 & 284 & 15.8\% \\
5 & Nnenum & 23 & 0 & 8 & 0 & 246 & 13.7\% \\
6 & Debona & 2 & 0 & 2 & 0 & 24 & 1.3\% \\
7 & Cgdtest & 2 & 0 & 2 & 2 & -176 & 0\% \\
\bottomrule
\end{tabular}
}
\end{center}
\end{table}

\begin{figure}[h]
\centerline{\includegraphics[width=\textwidth]{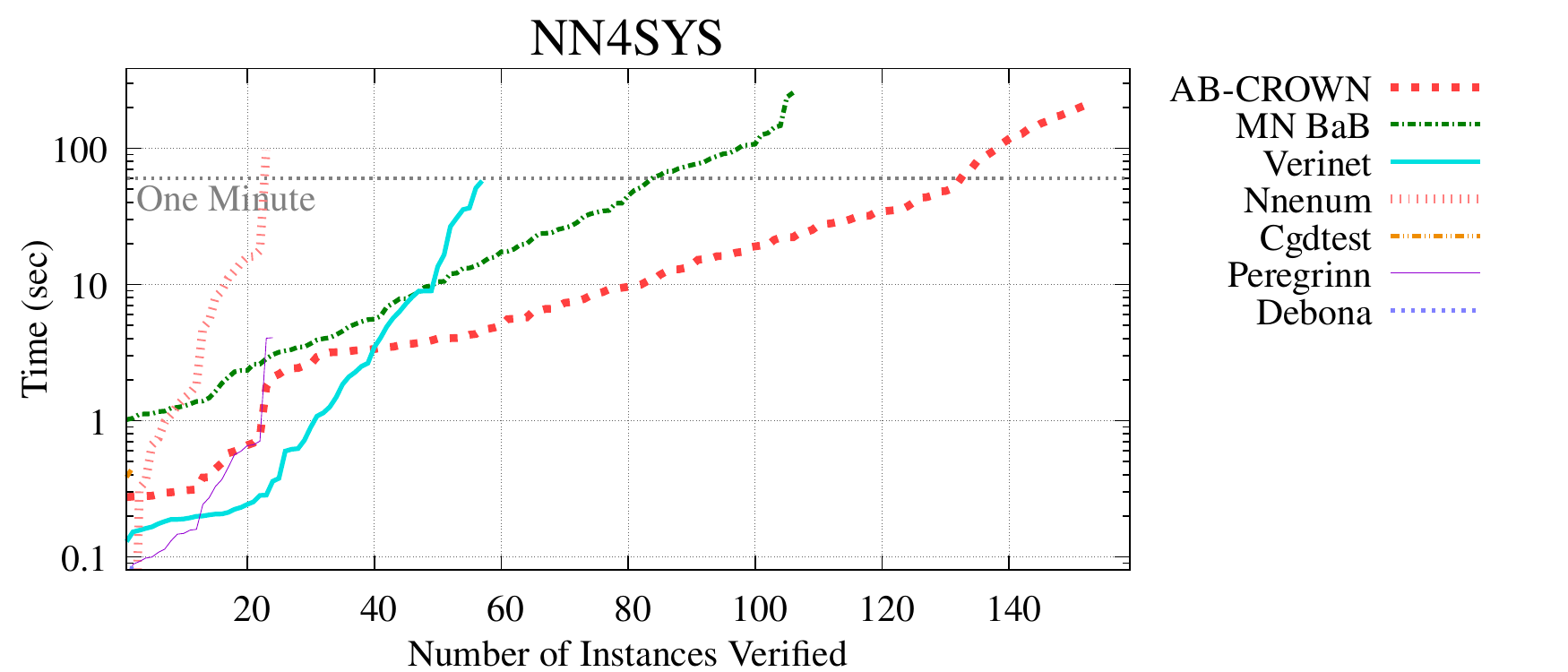}}
\caption{Cactus Plot for NN4SYS.}
\end{figure}


\begin{table}[h]
\begin{center}
\caption{Benchmark \texttt{oval21}} 
{\setlength{\tabcolsep}{2pt}
\begin{tabular}[h]{@{}llllllrr@{}}
\toprule
\textbf{\# ~} & \textbf{Tool} & \textbf{Verified} & \textbf{Falsified} & \textbf{Fastest} & \textbf{Penalty} & \textbf{Score} & \textbf{Percent}\\
\midrule
1 & $\alpha$,$\beta$ Crown & 25 & 1 & 10 & 0 & 291 & 100.0\% \\
2 & \textsc{MN-BaB} & 19 & 1 & 2 & 0 & 205 & 70.4\% \\
3 & Verinet & 17 & 1 & 1 & 0 & 189 & 64.9\% \\
4 & Marabou & 19 & 0 & 17 & 1 & 125 & 43.0\% \\
5 & Nnenum & 3 & 1 & 0 & 0 & 40 & 13.7\% \\
6 & Peregrinn & 1 & 0 & 0 & 0 & 10 & 3.4\% \\
7 & Cgdtest & 11 & 0 & 1 & 7 & -580 & 0\% \\
\bottomrule
\end{tabular}
}
\end{center}
\end{table}

\begin{figure}[h]
\centerline{\includegraphics[width=\textwidth]{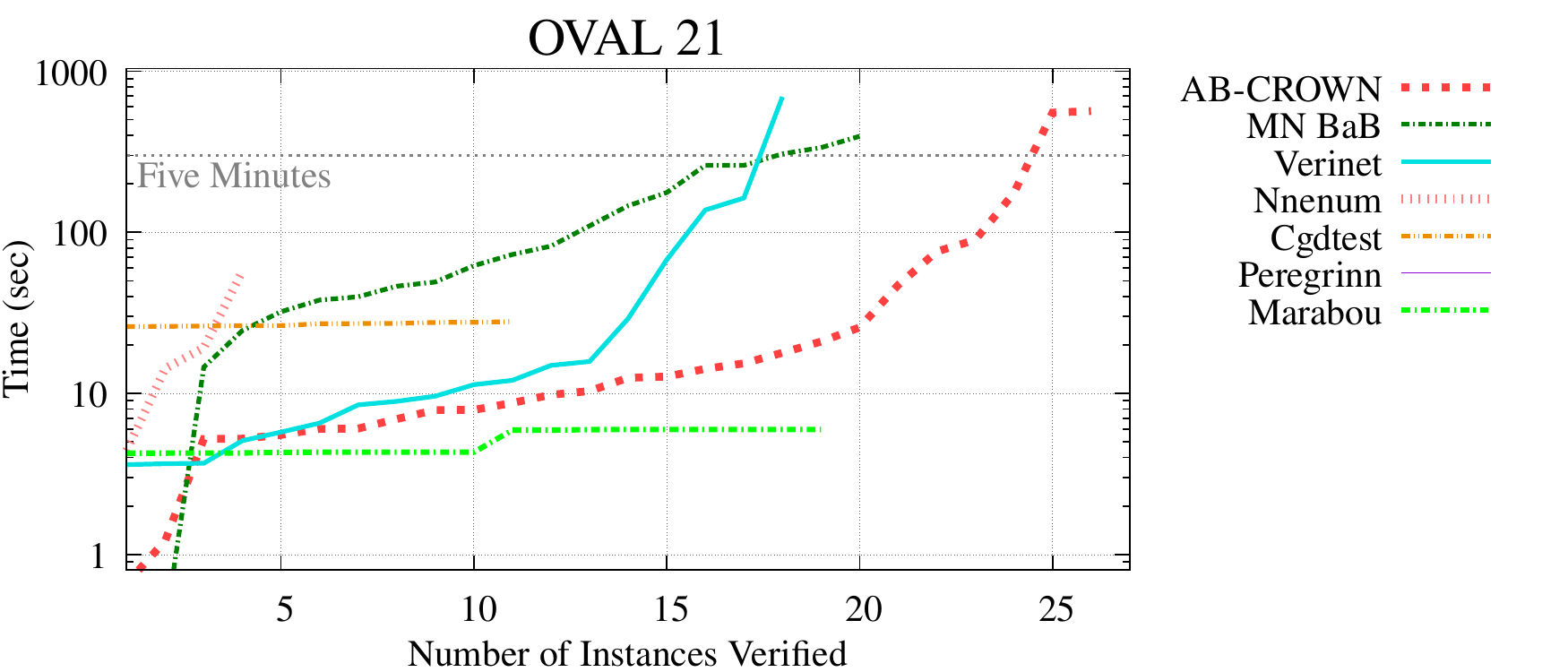}}
\caption{Cactus Plot for OVAL 21.}
\end{figure}


\begin{table}[h]
\begin{center}
\caption{Benchmark \texttt{reach-prob-density}} 
{\setlength{\tabcolsep}{2pt}
\begin{tabular}[h]{@{}llllllrr@{}}
\toprule
\textbf{\# ~} & \textbf{Tool} & \textbf{Verified} & \textbf{Falsified} & \textbf{Fastest} & \textbf{Penalty} & \textbf{Score} & \textbf{Percent}\\
\midrule
1 & Nnenum & 22 & 14 & 22 & 0 & 411 & 100.0\% \\
2 & $\alpha$,$\beta$ Crown & 22 & 14 & 23 & 0 & 406 & 98.8\% \\
3 & Verinet & 22 & 14 & 10 & 0 & 383 & 93.2\% \\
4 & \textsc{MN-BaB} & 22 & 12 & 14 & 0 & 368 & 89.5\% \\
5 & Marabou & 17 & 14 & 12 & 0 & 334 & 81.3\% \\
6 & Peregrinn & 18 & 14 & 2 & 0 & 324 & 78.8\% \\
7 & Cgdtest & 0 & 5 & 5 & 0 & 60 & 14.6\% \\
8 & Debona & 0 & 2 & 2 & 0 & 24 & 5.8\% \\
\bottomrule
\end{tabular}
}
\end{center}
\end{table}

\begin{figure}[h]
\centerline{\includegraphics[width=\textwidth]{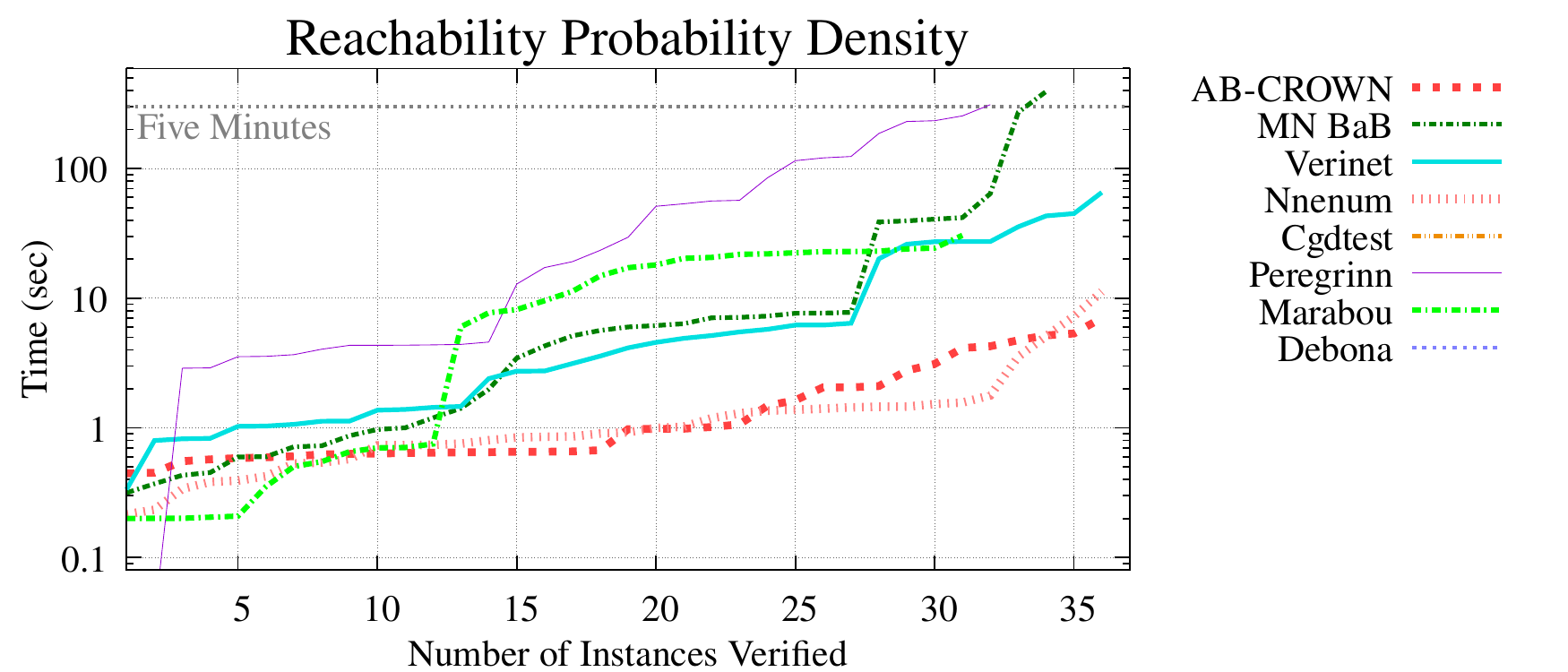}}
\caption{Cactus Plot for Reachability Probability Density.}
\end{figure}


\begin{table}[h]
\begin{center}
\caption{Benchmark \texttt{rl-benchmarks}} 
{\setlength{\tabcolsep}{2pt}
\begin{tabular}[h]{@{}llllllrr@{}}
\toprule
\textbf{\# ~} & \textbf{Tool} & \textbf{Verified} & \textbf{Falsified} & \textbf{Fastest} & \textbf{Penalty} & \textbf{Score} & \textbf{Percent}\\
\midrule
1 & $\alpha$,$\beta$ Crown & 193 & 103 & 296 & 0 & 3552 & 100.0\% \\
2 & Verinet & 193 & 103 & 292 & 0 & 3547 & 99.9\% \\
3 & \textsc{MN-BaB} & 193 & 103 & 288 & 0 & 3536 & 99.5\% \\
4 & Nnenum & 191 & 103 & 283 & 0 & 3506 & 98.7\% \\
5 & Peregrinn & 193 & 103 & 271 & 0 & 3502 & 98.6\% \\
6 & Marabou & 191 & 103 & 278 & 0 & 3496 & 98.4\% \\
7 & Debona & 153 & 99 & 240 & 0 & 3000 & 84.5\% \\
8 & Averinn & 92 & 8 & 16 & 0 & 1032 & 29.1\% \\
9 & Cgdtest & 10 & 24 & 29 & 3 & 98 & 2.8\% \\
10 & Verapak & 0 & 4 & 0 & 0 & 40 & 1.1\% \\
\bottomrule
\end{tabular}
}
\end{center}
\end{table}

\begin{figure}[h]
\centerline{\includegraphics[width=\textwidth]{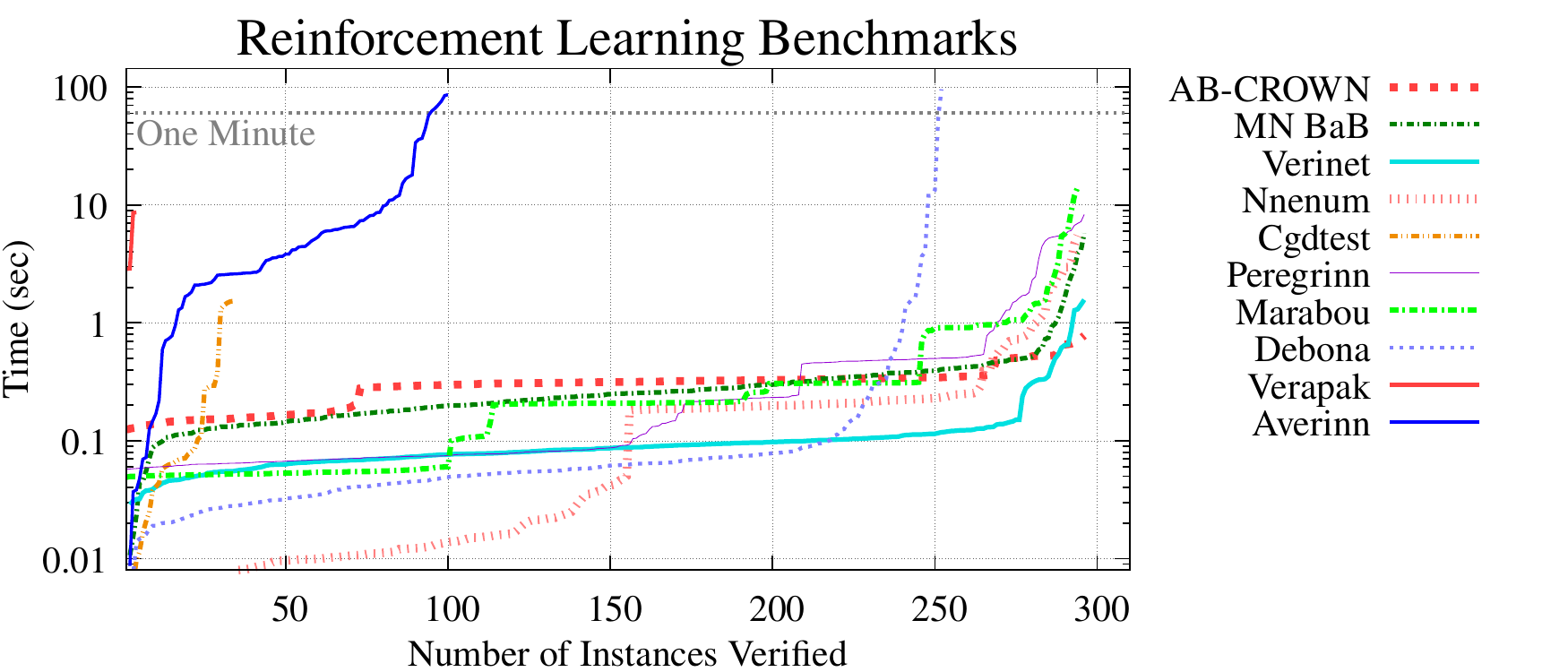}}
\caption{Cactus Plot for Reinforcement Learning Benchmarks.}
\end{figure}


\begin{table}[h]
\begin{center}
\caption{Benchmark \texttt{sri-resnet-a}} 
{\setlength{\tabcolsep}{2pt}
\begin{tabular}[h]{@{}llllllrr@{}}
\toprule
\textbf{\# ~} & \textbf{Tool} & \textbf{Verified} & \textbf{Falsified} & \textbf{Fastest} & \textbf{Penalty} & \textbf{Score} & \textbf{Percent}\\
\midrule
1 & $\alpha$,$\beta$ Crown & 20 & 12 & 7 & 0 & 356 & 100.0\% \\
2 & Cgdtest & 26 & 6 & 14 & 0 & 352 & 98.9\% \\
3 & \textsc{MN-BaB} & 18 & 12 & 19 & 0 & 343 & 96.3\% \\
4 & Verinet & 12 & 12 & 4 & 0 & 248 & 69.7\% \\
\bottomrule
\end{tabular}
}
\end{center}
\end{table}

\begin{figure}[h]
\centerline{\includegraphics[width=\textwidth]{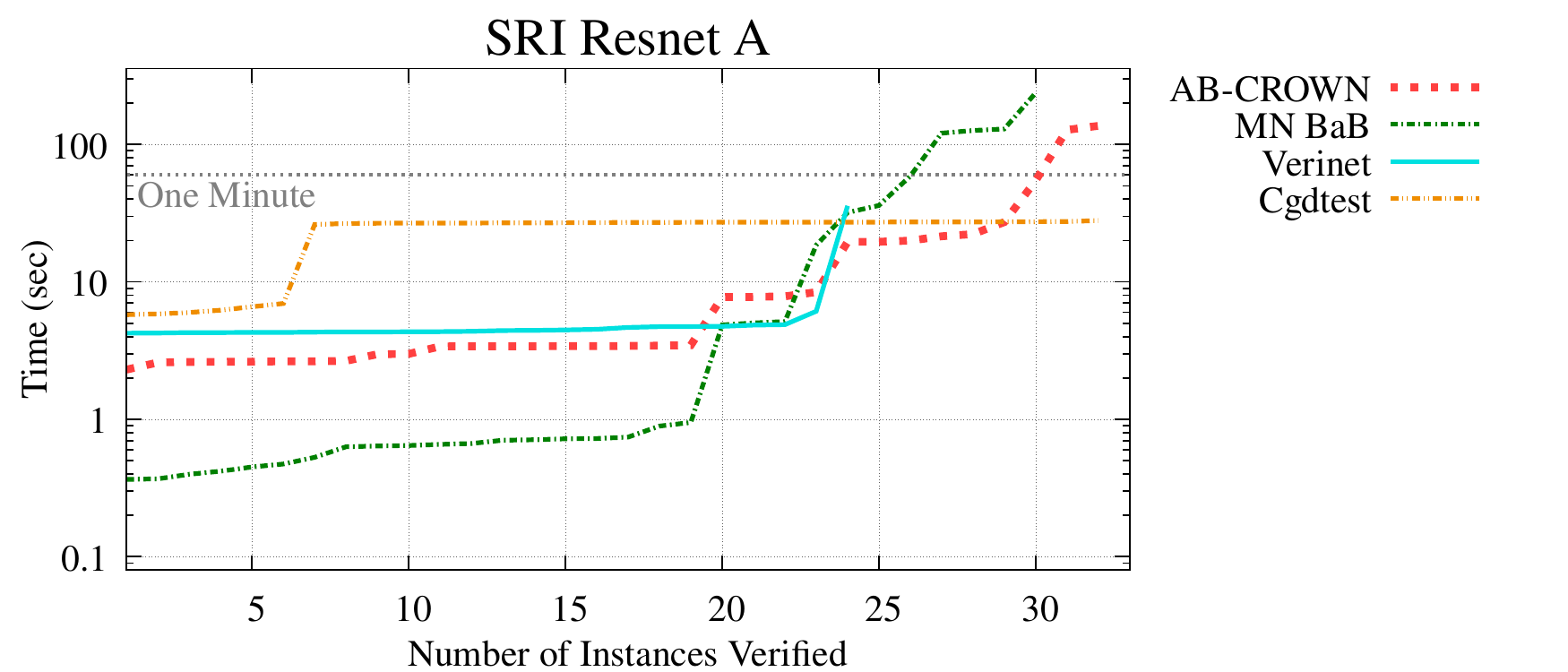}}
\caption{Cactus Plot for SRI Resnet A.}
\end{figure}


\begin{table}[h]
\begin{center}
\caption{Benchmark \texttt{sri-resnet-b}} 
{\setlength{\tabcolsep}{2pt}
\begin{tabular}[h]{@{}llllllrr@{}}
\toprule
\textbf{\# ~} & \textbf{Tool} & \textbf{Verified} & \textbf{Falsified} & \textbf{Fastest} & \textbf{Penalty} & \textbf{Score} & \textbf{Percent}\\
\midrule
1 & \textsc{MN-BaB} & 27 & 11 & 24 & 0 & 435 & 100.0\% \\
2 & $\alpha$,$\beta$ Crown & 28 & 11 & 9 & 0 & 435 & 100.0\% \\
3 & Cgdtest & 22 & 10 & 9 & 0 & 340 & 78.2\% \\
4 & Verinet & 20 & 11 & 4 & 0 & 321 & 73.8\% \\
\bottomrule
\end{tabular}
}
\end{center}
\end{table}

\begin{figure}[h]
\centerline{\includegraphics[width=\textwidth]{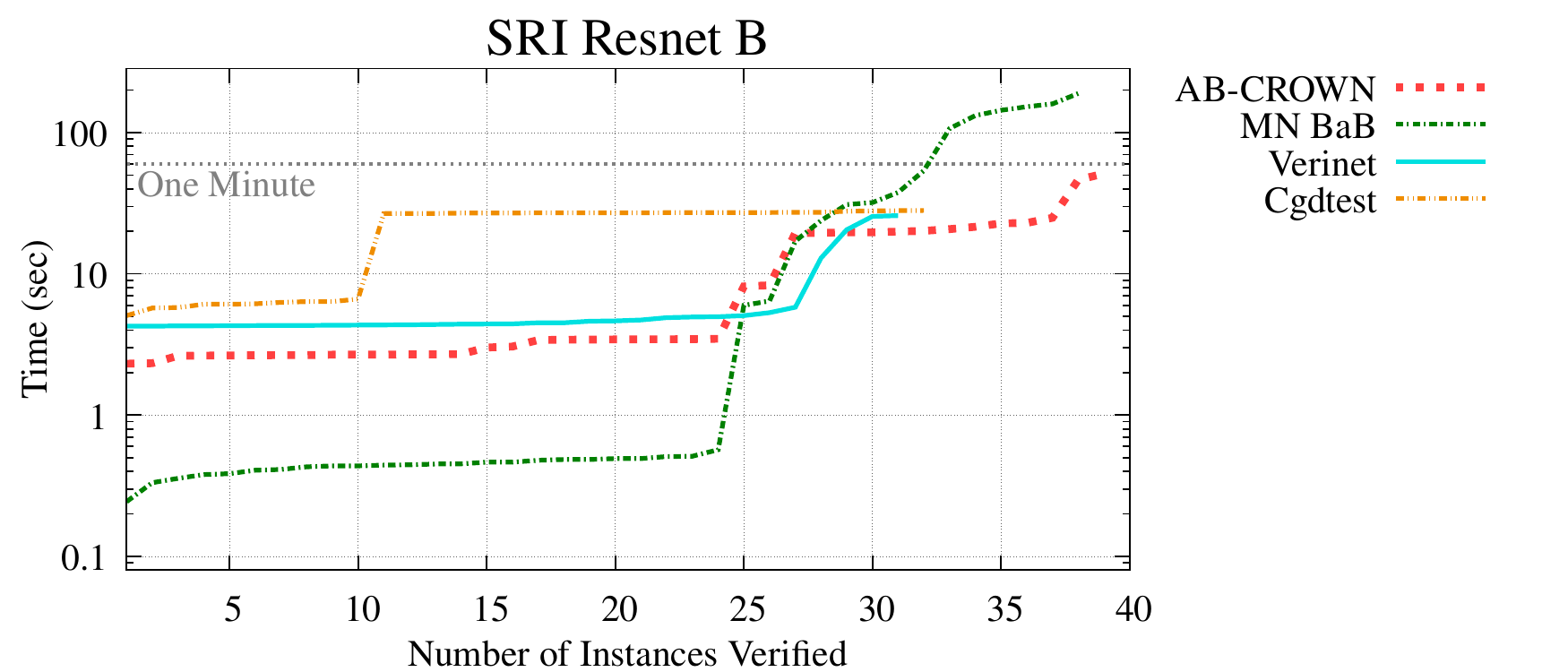}}
\caption{Cactus Plot for SRI Resnet B.}
\end{figure}


\begin{table}[h]
\begin{center}
\caption{Benchmark \texttt{tllverifybench}} 
{\setlength{\tabcolsep}{2pt}
\begin{tabular}[h]{@{}llllllrr@{}}
\toprule
\textbf{\# ~} & \textbf{Tool} & \textbf{Verified} & \textbf{Falsified} & \textbf{Fastest} & \textbf{Penalty} & \textbf{Score} & \textbf{Percent}\\
\midrule
1 & Fastbatllnn & 11 & 21 & 32 & 0 & 384 & 100.0\% \\
2 & \textsc{MN-BaB} & 11 & 21 & 21 & 0 & 364 & 94.8\% \\
3 & $\alpha$,$\beta$ Crown & 11 & 21 & 12 & 0 & 353 & 91.9\% \\
4 & Peregrinn & 10 & 21 & 7 & 0 & 324 & 84.4\% \\
5 & Verinet & 11 & 21 & 0 & 0 & 320 & 83.3\% \\
6 & Nnenum & 1 & 21 & 10 & 0 & 240 & 62.5\% \\
7 & Debona & 0 & 19 & 10 & 0 & 210 & 54.7\% \\
8 & Marabou & 4 & 15 & 2 & 0 & 194 & 50.5\% \\
9 & Cgdtest & 0 & 9 & 6 & 1 & 2 & 0.5\% \\
\bottomrule
\end{tabular}
}
\end{center}
\end{table}

\begin{figure}[h]
\centerline{\includegraphics[width=\textwidth]{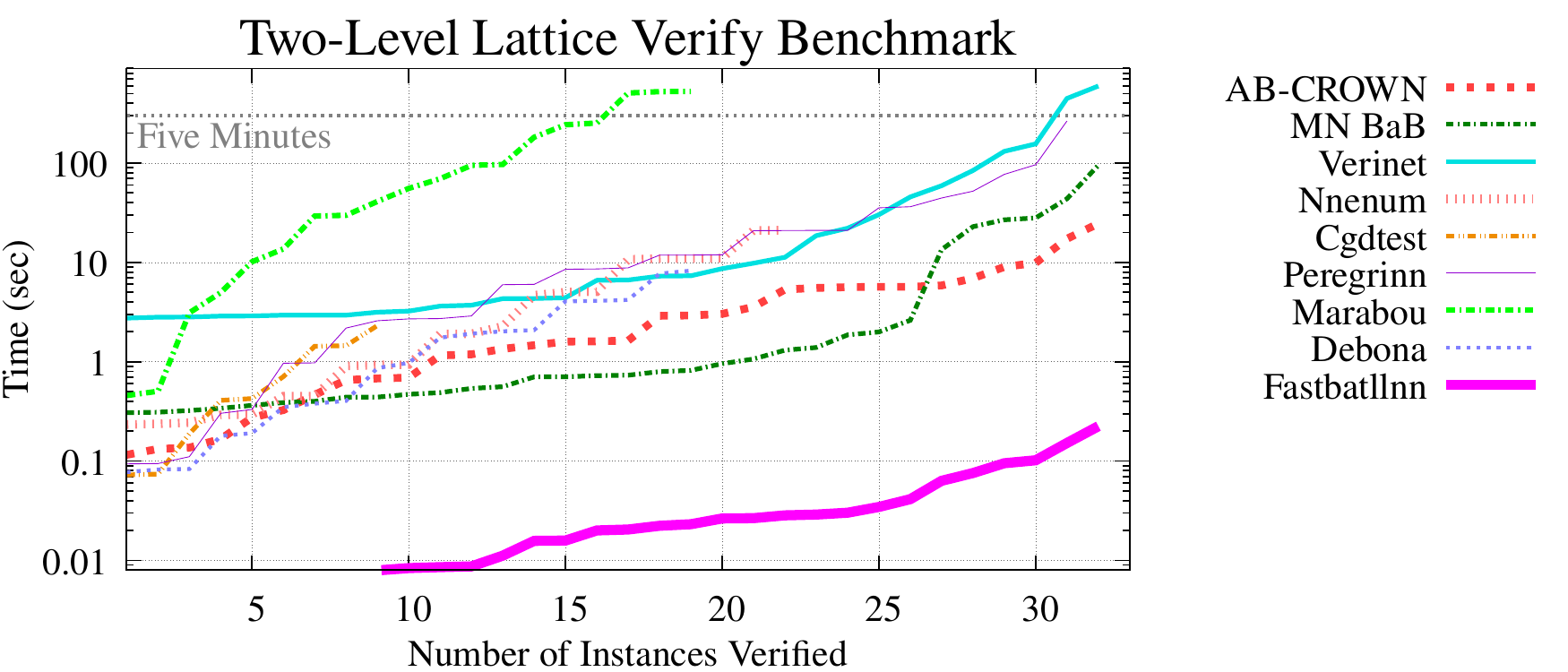}}
\caption{Cactus Plot for Two-Level Lattice Verify Benchmark.}
\end{figure}


\begin{table}[h]
\begin{center}
\caption{Benchmark \texttt{vggnet16-2022}} 
{\setlength{\tabcolsep}{2pt}
\begin{tabular}[h]{@{}llllllrr@{}}
\toprule
\textbf{\# ~} & \textbf{Tool} & \textbf{Verified} & \textbf{Falsified} & \textbf{Fastest} & \textbf{Penalty} & \textbf{Score} & \textbf{Percent}\\
\midrule
1 & $\alpha$,$\beta$ Crown & 14 & 1 & 11 & 0 & 176 & 100.0\% \\
2 & Nnenum & 11 & 1 & 0 & 0 & 127 & 72.2\% \\
3 & \textsc{MN-BaB} & 5 & 1 & 4 & 0 & 69 & 39.2\% \\
4 & Verinet & 5 & 1 & 0 & 0 & 60 & 34.1\% \\
5 & Cgdtest & 0 & 2 & 1 & 4 & -378 & 0\% \\
\bottomrule
\end{tabular}
}
\end{center}
\end{table}

\begin{figure}[h]
\centerline{\includegraphics[width=\textwidth]{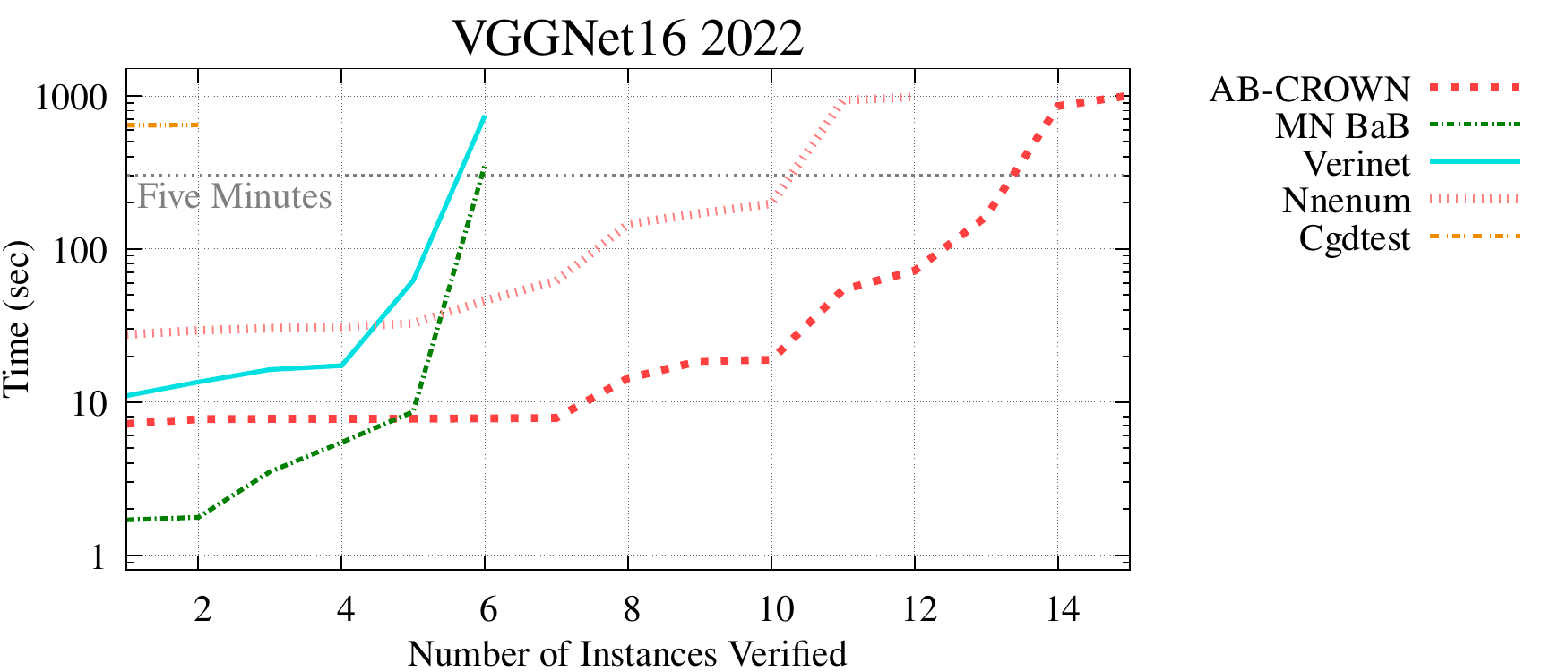}}
\caption{Cactus Plot for VGGNet16 2022.}
\end{figure}

\clearpage
\subsection{Unscored Benchmarks}
\label{sec:benchmark_results_unscored}


\begin{table}[h]
\begin{center}
\caption{Benchmark \texttt{acasxu}} 
{\setlength{\tabcolsep}{2pt}
\begin{tabular}[h]{@{}llllllrr@{}}
\toprule
\textbf{\# ~} & \textbf{Tool} & \textbf{Verified} & \textbf{Falsified} & \textbf{Fastest} & \textbf{Penalty} & \textbf{Score} & \textbf{Percent}\\
\midrule
1 & Nnenum & 139 & 47 & 174 & 0 & 2218 & 100.0\% \\
2 & $\alpha$,$\beta$ Crown & 139 & 46 & 59 & 0 & 2021 & 91.1\% \\
3 & \textsc{MN-BaB} & 110 & 46 & 52 & 0 & 1664 & 75.0\% \\
4 & Cgdtest & 85 & 30 & 115 & 7 & 680 & 30.7\% \\
\bottomrule
\end{tabular}
}
\end{center}
\end{table}

\begin{figure}[h]
\centerline{\includegraphics[width=\textwidth]{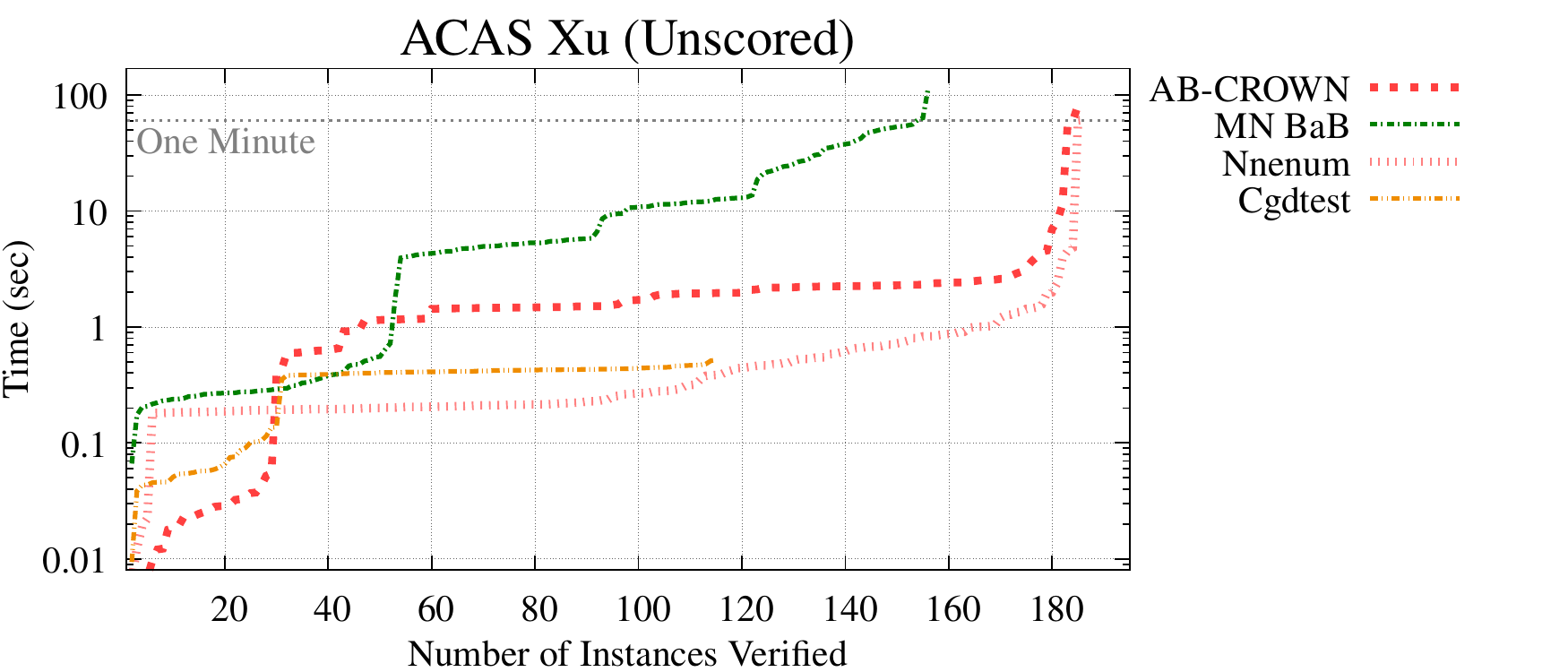}}
\caption{Cactus Plot for ACAS Xu (Unscored).}
\end{figure}


\begin{table}[h]
\begin{center}
\caption{Benchmark \texttt{cifar2020}} 
{\setlength{\tabcolsep}{2pt}
\begin{tabular}[h]{@{}llllllrr@{}}
\toprule
\textbf{\# ~} & \textbf{Tool} & \textbf{Verified} & \textbf{Falsified} & \textbf{Fastest} & \textbf{Penalty} & \textbf{Score} & \textbf{Percent}\\
\midrule
1 & Verinet & 91 & 35 & 109 & 0 & 1486 & 100.0\% \\
2 & $\alpha$,$\beta$ Crown & 95 & 34 & 78 & 0 & 1479 & 99.5\% \\
3 & \textsc{MN-BaB} & 93 & 28 & 26 & 0 & 1275 & 85.8\% \\
4 & Nnenum & 66 & 19 & 0 & 0 & 850 & 57.2\% \\
5 & Cgdtest & 63 & 26 & 5 & 6 & 305 & 20.5\% \\
6 & Verapak & 0 & 15 & 1 & 0 & 152 & 10.2\% \\
7 & Marabou & 4 & 0 & 0 & 1 & -60 & 0\% \\
\bottomrule
\end{tabular}
}
\end{center}
\end{table}

\begin{figure}[h]
\centerline{\includegraphics[width=\textwidth]{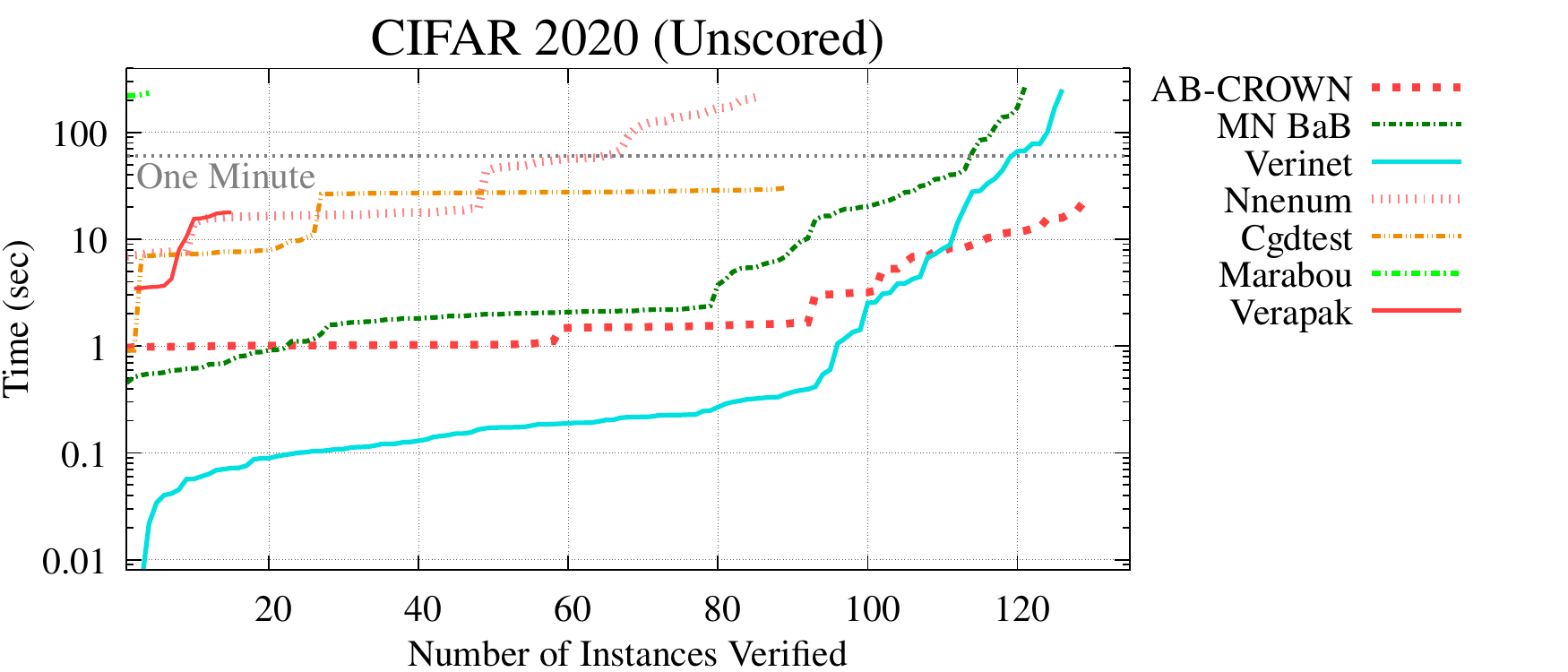}}
\caption{Cactus Plot for CIFAR 2020 (Unscored).}
\end{figure}

\clearpage
\subsection{Detailed Results}
\label{sec:results_detailed}

\begin{center}
{\setlength{\tabcolsep}{1pt}
\scriptsize
\begin{longtable}{@{}llllllllllllll@{}}
\caption{\footnotesize Instance Runtimes. Fastest times are \textcolor{blue}{blue}. Second fastest are \textcolor{second}{green}. Penalties are red crosses (\textbf{\textcolor{red}{\ding{55}}}).} \label{tab:all_results} \\
\toprule
\textbf{Category} & \textbf{Id} & \textbf{Result} & \textbf{$\alpha$,$\beta$-C} & \textbf{MnB} & \textbf{Verin} & \textbf{nnen} & \textbf{CGD} & \textbf{Pereg} & \textbf{Marab} & \textbf{Debon} & \textbf{FastBaT} & \textbf{Verap} & \textbf{Averi} \\
\midrule
\endhead
Carvana 2022 & 0 & ~\textsc{unsat} & \textcolor{blue}{7.53} & - & - & - & - & - & - & - & - & - & - \\
Carvana 2022 & 1 & ~\textsc{unsat} & \textcolor{blue}{26.1} & \textcolor{second}{76.4} & - & - & - & - & - & - & - & - & - \\
Carvana 2022 & 2 & ~? & - & - & - & - & - & - & - & - & - & - & - \\
Carvana 2022 & 3 & ~? & - & - & - & - & - & - & - & - & - & - & - \\
Carvana 2022 & 4 & ~\textsc{unsat} & \textcolor{blue}{7.83} & - & - & - & - & - & - & - & - & - & - \\
Carvana 2022 & 5 & ~\textsc{unsat} & \textcolor{blue}{8.73} & - & - & - & - & - & - & - & - & - & - \\
Carvana 2022 & 6 & ~\textsc{unsat} & \textcolor{blue}{26.9} & - & - & - & - & - & - & - & - & - & - \\
Carvana 2022 & 7 & ~? & - & - & - & - & - & - & - & - & - & - & - \\
\bottomrule
\end{longtable}
}
\end{center}


\end{document}